\documentclass[lettersize,journal]{IEEEtran}

\usepackage[utf8]{inputenc}
\usepackage[T1]{fontenc}
\usepackage[caption=false,font=normalsize,labelfont=sf,textfont=sf]{subfig}

\usepackage{url}           
\usepackage{booktabs}       
\usepackage{amsfonts}       
\usepackage{nicefrac}       
\usepackage{microtype}      
\usepackage{xcolor}
\usepackage{graphicx}
\usepackage{amsmath}
\usepackage{lipsum}
\usepackage{wrapfig}
\usepackage{adjustbox}
\usepackage{multirow}
\usepackage{paralist}
\usepackage{CJKutf8}
\usepackage{subcaption}
\usepackage{colortbl}
\usepackage[textsize=tiny]{todonotes}
\usepackage{arydshln}
\usepackage[most]{tcolorbox}
\usepackage{enumitem}

\definecolor{newgreen}{rgb}{0.7, 0.9, 0.7}
\definecolor{newblue}{rgb}{0.85, 0.85, 0.9}
\newcommand{\CFRB}{\texttt{CFRB}}
\newcommand{\CIRB}{\texttt{CIRB}}
\newcommand{\PFRB}{\texttt{PFRB}}

\definecolor{tkcolor}{RGB}{224,223,255}
\definecolor{shallowred}{RGB}{242,204,208}
\newtcolorbox{takeaways}[1][]{
	width=\columnwidth,
	colback = tkcolor, 
	colframe = tkcolor, 
	boxsep=0pt,left=10pt,right=10pt,top=5pt,bottom=5pt,
	fontupper=\linespread{0.9}\selectfont,
	title=#1}

\newtcolorbox{limitation}[1][]{
	width=\columnwidth,
	colback = shallowred, 
	colframe = shallowred, 
	boxsep=0pt,left=10pt,right=10pt,top=5pt,bottom=5pt,
	fontupper=\linespread{0.9}\selectfont,
	title=#1}
\definecolor{shallowgray}{RGB}{237,240,246}
\newtcolorbox{prompt}[1][]{
	width=\columnwidth,
	colback = shallowgray, 
	colframe = shallowgray, 
	boxsep=0pt,left=10pt,right=10pt,top=5pt,bottom=5pt,
	fontupper=\linespread{0.9}\selectfont,
	title=#1}
\newtheorem{theorem}{Theorem}

\newtheorem{definition}[theorem]{Definition}
\newtheorem{assumption}[theorem]{Assumption}

\usepackage{hyperref}      
\title{RBF++: Quantifying and Optimizing Reasoning Boundaries across Measurable and Unmeasurable Capabilities for Chain-of-Thought Reasoning}

\author{
	Qiguang Chen, Libo Qin, Jinhao Liu, Yue Liao, Jiaqi Wang, Jingxuan Zhou, Wanxiang Che\\
	\thanks{Qiguang Chen, Jinhao Liu and Wanxiang Che are with Research Center for Social Computing and Interactive Robotics, Harbin Institute of Technology, Harbin, China, and also with Faculty of Computing, Harbin Institute of Technology, Harbin, China. Email: \{qgchen, car\}@ir.hit.edu.cn.}
	\thanks{Libo Qin and Jingxuan Zhou are with School of Computer Science and Engineering, Central South University, Changsha, China. Email: lbqin@csu.edu.cn.}
	\thanks{Jiaqi Wang are with Chinese University of Hong Kong, Hong Kong, China. Email: jqwang23@cse.cuhk.edu.hk.}
	\thanks{Yue Liao are with MMLab, The Chinese University of Hong Kong, Hong Kong, China. Email: liaoyue.ai@gmail.com.}
	\thanks{The corresponding author is Wanxiang Che and Libo Qin.}
	\thanks{A preliminary version of this research has appeared in NeurIPS 2024 (Oral)~\cite{chen2024unlocking}.}
}
\begin{document}

	\maketitle

	\begin{abstract}
		Chain-of-Thought (CoT) reasoning has proven effective in enhancing large language models (LLMs) on complex tasks, spurring research into its underlying mechanisms. However, two primary challenges remain for real-world applications:  (1) \textit{the lack of quantitative metrics and actionable guidelines for evaluating and optimizing measurable boundaries of CoT capability}, and (2) \textit{the absence of methods to assess boundaries of unmeasurable CoT capability, such as multimodal perception}.
		To address these gaps, we introduce the Reasoning Boundary Framework++ (RBF++).
		To tackle the first challenge, we define the reasoning boundary (RB) as the maximum limit of CoT performance. We also propose a combination law for RBs, enabling quantitative analysis and offering actionable guidance across various CoT tasks.
		For the second challenge, particularly in multimodal scenarios, we introduce a constant assumption, which replaces unmeasurable RBs with scenario-specific constants. Additionally, we propose the reasoning boundary division mechanism, which divides unmeasurable RBs into two sub-boundaries, facilitating the quantification and optimization of both unmeasurable domain knowledge and multimodal perception capabilities.
		Extensive experiments involving 38 models across 13 tasks validate the feasibility of our framework in cross-modal settings. Additionally, we evaluate 10 CoT strategies, offer insights into optimization and decay from two complementary perspectives, and expand evaluation benchmarks for measuring RBs in LLM reasoning.
		We hope this work advances the understanding of RBs and optimization strategies in LLMs. Code and data are available at \url{https://github.com/LightChen233/reasoning-boundary}.
	\end{abstract}
\begin{IEEEkeywords}
    Reasoning Boundary, Multimodal Reasoning, Chain of Thought, Reasoning Behavior Explanation.
\end{IEEEkeywords}
	\section{Introduction}
\IEEEPARstart{R}{ecently}, Large Language Models (LLMs) have made notable strides, demonstrating an expanding range of capabilities and applications across various tasks~\cite{zhao2023survey,pan2023preliminary,qin2024large}. Among the most advanced LLMs, the GPT~\cite{brown2020language,openai2022gpt35,jaech2024openai}, and DeepSeek~\cite{liu2024deepseek,guo2025deepseek} series stand out for their emergent abilities. A prominent example of such capabilities is the Chain-of-Thought (CoT) reasoning~\cite{nye2022show,wei2022chain,lin2025navcot}, which enables models to articulate their reasoning step-by-step. This methodology enhances the models' prediction accuracy by anchoring their decision-making processes in structured, logical reasoning~\cite{wei2022chain,kojima2022large,hu2023tree,qin2023cross,fu2025reason,chen-etal-2024-m3cot}.

\begin{figure*}[t]
	\centering
	\includegraphics[width=0.95\textwidth]{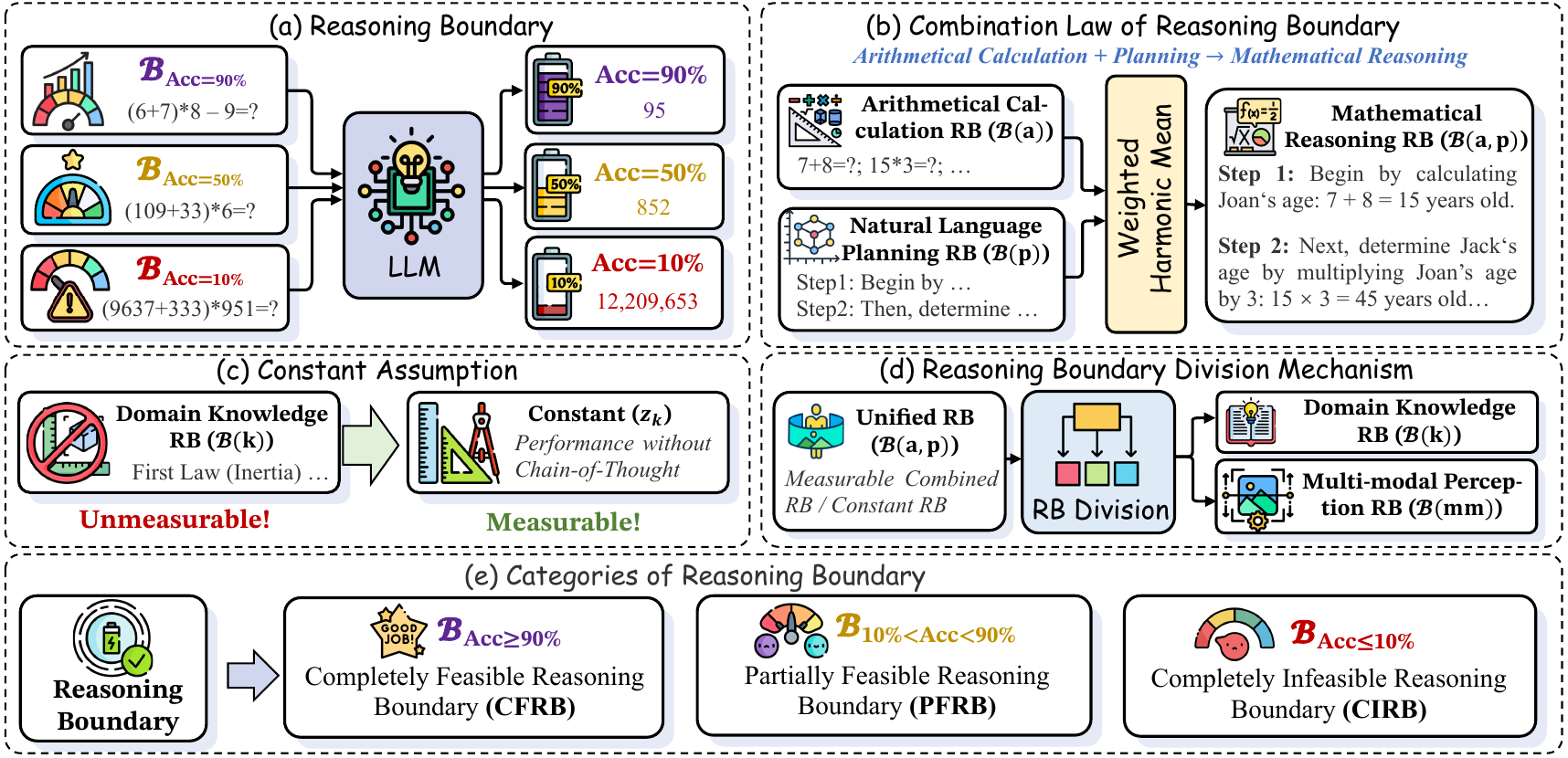}
	\caption{Overview of the introduced concepts: (a) reasoning boundary (RB), (b) combination law for quantifying the upper bound of LLM capabilities in measurable scenarios; (c) constant assumption and (d) RB division mechanism for unmeasurable scenarios; (e) categories of RB for reasoning optimization guidance.}
	\label{fig:intro}
\end{figure*}

Recent studies have explored the mechanisms underlying CoT to better understand its operational dynamics and improve its predictability and controllability. In this regard, Madaan \textit{et al.}~\cite{madaan-etal-2023-makes} and Wang \textit{et al.}~\cite{wang-etal-2023-towards} first conduct a qualitative boundary analysis, demonstrating that CoT is constrained by the reasoning logic of contextual demonstrations. Extending this line of research, Bi \textit{et al.}~\cite{bi2024program} investigate these boundaries in code planning tasks by training LLMs on CoT samples with varying complexity levels, revealing that model performance deteriorates beyond a certain complexity threshold. To further characterize CoT's limitations, Feng \textit{et al.}~\cite{feng2024towards} propose a theoretical framework of single-step computation, suggesting that model performance is upper-bounded by the input length in single-step reasoning.
Despite significant progress in this field, the precise boundaries of CoT and the mechanisms by which these boundaries affect CoT performance remain unclear. Current studies encounter two main challenges: (1) \textbf{Lack of quantitative metrics and actionable guidance for evaluating and optimizing boundaries of measurable CoT capability: } Much of the existing research relies on qualitative assessments, as depicted in Fig.~\ref{fig:intro} (a, b), resulting in a lack of standardized, quantitative metrics to guide the optimization of CoT for measurable capabilities such as calculation numbers and planning steps. This gap limits the practical application of CoT research.
(2) \textbf{Absence of methods to assess boundaries of unmeasurable CoT capability: } As shown in Fig.~\ref{fig:intro} (c, d), boundaries in complex scenarios involve unmeasurable capabilities (e.g., multimodal perception and multi-domain knowledge) that cannot be quantified, preventing the optimization of corresponding CoT RBs in such tasks.

To address these challenges, we introduce novel \textit{reasoning boundary framework++} (RBF++) to systematically examine and optimize the reasoning boundaries of current LLMs.
(1) To address the first challenge, we formally define a model’s reasoning boundary (RB) as its ability to achieve a target accuracy under a given measurable task difficulty. To extend this to practical applications, we introduce a \textit{combination law} that generalizes RBs for complex, real-world scenarios. Further, to optimize CoT reasoning, we identify and analyze three distinct RB intervals. These intervals guide the refinement of RBs and the optimization of reasoning paths under the combination law, leading to superior performance on our proposed benchmark.
(2) To address the second challenge in complex tasks, particularly in multi-modal scenarios, we propose a constant assumption, enabling the quantification of boundaries for otherwise unmeasurable  capabilities, such as the multimodal multi-domain CoT RB. More practically, we introduce a boundary division mechanism to measure the fine-grained unmeasurable RBs, which refine coarse-grained RBs (e.g., unified multimodal multi-domain CoT RB) into more fine-grained RBs (e.g., multimodal perception RB and domain knowledge RB), thereby enhancing the  adaptability across more diverse domains and modalities.

Moreover, inspired by advanced reasoning LLMs~\cite{chen2025towards}, we propose the BigGSM++ benchmark to evaluate the RBs of advanced reasoning LLMs such as DeepSeek-R1~\cite{guo2025deepseek}. Our findings show that Reinforcement Learning substantially enhances these RBs, achieving a 100x improvement in previously infeasible RBs with 10\% accuracy, while minimally affecting RBs at 90\% accuracy.
We validate RBF++ across 38 models and 13 CoT tasks, demonstrating its ability to generalize in both measurable and unmeasurable domains, including multimodal commonsense, scientific knowledge, and multi-modal perception.
Additionally, we clarify the effectiveness and limitations of CoT optimization strategies. For example, we explain why Program-of-Thought, Least-to-Most, Complex-CoT succeed in textual scenarios but fail in multimodal ones. Finally, we introduce the Minimum Acceptable Reasoning Path (MARP) and MARP++, which outperforms existing methods at least 2\% accuracy both in textual and multi-modal reasoning.

Our key contributions in this work are as follows: (1) We introduce RBF++ to quantify the boundaries of both measurable and unmeasurable CoT capabilities, guiding the optimization of CoT strategies; (2) We analyze the effectiveness and limitations of CoT optimization strategies across 10 CoT strategies, 38 models and 13 CoT tasks; (3) We present the MARP and MARP++ prompting methods, which significantly enhance textual and multimodal reasoning performance.

This paper extends our previous work, RBF~\cite{chen2024unlocking} (conference version), by significantly enhancing several key aspects: (1) \textit{Deeper Advanced Reasoning Model Exploration}: The BigGSM++ benchmark is introduced, enabling the investigation of  RBs of advanced reasoning LLMs, such as DeepSeek-R1. This uncovers growth patterns in their reasoning capabilities and provides valuable development insights. (2) \textit{Broader Cross-Modal Application Expansion}: We extend the framework that incorporates a scenario-dependent constant assumption and reasoning boundary division mechanism, enabling wider applicability across multi-modal scenarios with unmeasurable capabilities.
(3) \textit{Comprehensive Experimental Evaluation}: This work includes additional analyses, offering more extensive evaluations than the conference version, which provides a deeper and more thorough understanding of the effectiveness and  limitations of CoT strategies in multimodal settings.
(4) \textit{Optimized Cross-modal Prompting for Enhanced Reasoning}: We introduce the MARP++ strategy, improving multi-domain multimodal reasoning accuracy by over 5\%.

\section{Reasoning Boundary Framework++ (RBF++)}
We present the Reasoning Boundary Framework++ (RBF++), which includes five key components: (1) a reasoning boundary (RB) to define the limits of reasoning capabilities; (2) a combination law to quantify the combined RB of multiple sub-RBs; (3) constant assumptions to define the RB of unmeasurable CoT capabilities; (4) a reasoning boundary division mechanism for more detailed sub-RBs of unmeasurable CoT capabilities; and (5) categories of RBs to guide CoT optimization.
\subsection{Reasoning Boundary}
\label{sec:definition}
To formally measure the reasoning capacity of models, as shown in Figure~\ref{fig:intro} (a), we introduce the concept of reasoning boundary (RB), which defines the maximum problem difficulty at which an LLM maintains acceptable performance. RB represents the point beyond which the model’s accuracy drops below a target threshold. Formally, for a model $m$ on a task $t$, RB is the greatest difficulty level $d$ at which the model’s accuracy exceeds a predefined threshold $K_1$:
\begin{equation}
	\label{eq:max_available}
\mathcal{B}_{Acc=K_1}(t|m) = \sup_{d} \{ d | Acc(t|d, m) \le K_1 \},
\end{equation}
where $Acc(t|d, m)$ denote the accuracy of the LLMs on task $t$ given a difficulty level $d$. Difficulty can be defined as factors such as the number of reasoning steps, local operations or computational complexity.

\subsection{Combination Law of Reasoning Boundary}
In practical applications, LLMs typically integrate multiple capabilities to solve tasks effectively. To quantify the cooperation via the CoT mechanism, we propose the ``\textit{Combination Law of RB}'', which estimates the combined RB $\mathcal{B}_{\text{Acc}=K_1}(t_1, t_2, \dots, t_n|m)$ for $n$ tasks within LLM $m$ as follows:
\begin{equation}
	\mathcal{B}_{\text{Acc}=K_1}(t_1, t_2, \dots, t_n|m) \approx \frac{1}{\sum^{n}_{i=1}\frac{N_{i}}{\mathcal{B}_{\text{Acc}=K_1}(t_i|m)-b_i }}, \label{eq:combine-law}
\end{equation}
where $\mathcal{B}_{\text{Acc}=K_1}(t_i|m)$ represents the RB of model $m$ for task $t_i$.
$N_{i}$, and $b_i$ are task-specific scaling factors.
Further, in order to express the formula more concisely, we concisely represent normalized RB $\frac{\mathcal{B}_{\text{Acc}=K_1}(t_i|m)-b_i}{N_i}$ as $\mathcal{B}(t_i)$. Therefore, the combined reasoning boundary can be expressed as:
\begin{equation}
	\mathcal{B}(t_1, t_2, \dots, t_n) \approx \frac{1}{\sum^{n}_{i=1}\frac{1}{\mathcal{B}(t_i)}}. \label{eq:combine-law-simple}
\end{equation}
As shown in Figure~\ref{fig:intro} (b), Eq.~\eqref{eq:combine-law-simple} provides a formula to estimate the combined RBs from the individual ones, offering insights into model behavior for complex tasks. See Appendix A for more mathematical analysis.

\subsection{Constant Assumptions}
To quantify the unmeasurable CoT capabilities, as shown in Figure~\ref{fig:intro} (c), we introduce a domain-specific constant to model's RBs of unmeasurable capabilities.
Formally, when some sub-boundaries are hard to measure but others remain stable, we evaluate the top-$j$ measurable sub-RBs, denoted as $\{B_i\}^{j}_{i=1}$. For unmeasurable sub-RBs beyond this range, we introduce constants $Z=\{z_i\}^{N}_{i=j+1}$ to replace the stable but unmeasurable sub-RBs $\{B_i\}^{N}_{i=j+1}$, which can be expressed as:
\begin{equation}
	\mathcal{B}(t_1, t_2, \dots, t_n) \approx \frac{1}{\sum^{j}_{i=1}\frac{N_{i}}{\mathcal{B}(t_i)-b_i }+\sum^{N}_{i=j+1}z_i}. \label{eq:constant-assumption}
\end{equation}
This approach approximates the reasoning boundary by combining measurable boundary contributions with constant values $z_i$ for unmeasurable boundaries.
Additionally, the framework allows the neglect of unmeasurable or incompletely divided boundaries, as they only affect the value of $Z$ without significantly influencing the overall analysis.

\subsection{Reasoning Boundary Division Mechanism} 
Due to the weighted harmonic mean in combination law, it offers advantageous mathematical properties. It allows us to divide a unified RB into more independent RBs, thereby enhancing its acceptability to address broader problems.
Specifically, given a unified RBs $B(p,o,v)$, it includes planning RB $B(p)$, operation RB $B(o)$, vertical domain RB $B(v)$, following:
\begin{equation}
	\mathcal{B}(p,o,v) = \frac{1}{\frac{1}{\mathcal{B}(p)}+\frac{1}{\mathcal{B}(o)}+\frac{1}{\mathcal{B}(v)}}.
\end{equation}
Further division of the vertical domain boundary $\mathcal{B}_v$ into $\mathcal{B}_{v1}$ and $\mathcal{B}_{v2}$ is straightforward, as the following relationship holds:
\begin{equation}
	\mathcal{B}(v) = \frac{1}{\frac{1}{\mathcal{B}(v_1)} + \frac{1}{\mathcal{B}(v_2)}}.
\end{equation}
Hence, as shown in Figure~\ref{fig:intro} (d), the overall boundary equation can be extended as follows:
\begin{equation}
	\mathcal{B}(p,o,v) = \frac{1}{\frac{1}{\mathcal{B}(p)}+\frac{1}{\mathcal{B}(o)}+\frac{1}{\mathcal{B}(v_1)} + \frac{1}{\mathcal{B}(v_2)}}.
\end{equation}
This formulation facilitates flexible and systematic boundary division across measurable and unmeasurable scenarios, thereby increasing the framework's practical utility in diverse domains.

\begin{figure*}[t]
	\centering
	\includegraphics[width=0.99\textwidth]{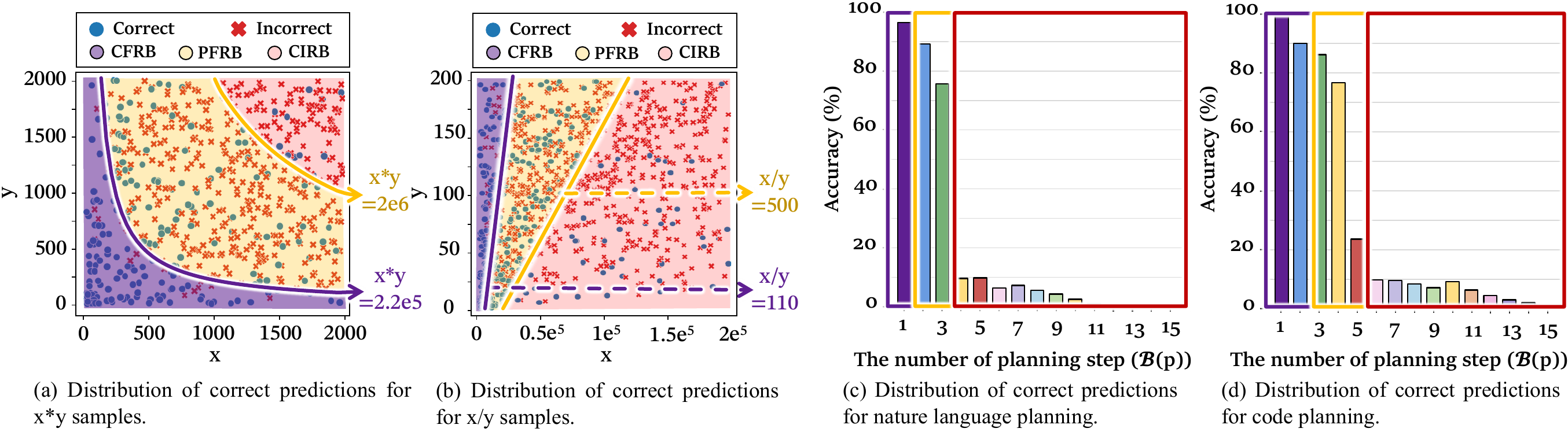}
	\caption{Existence Verification for Reasoning Boundary. Figures (b, c) present evaluations performed on BigGSM, where the reasoning accuracy of each step is manually analyzed, without considering whether the final conclusions are correct. }
	\label{fig:atom-rg}
\end{figure*}
\begin{figure*}[t]
	\centering
	\includegraphics[width=0.99\textwidth]{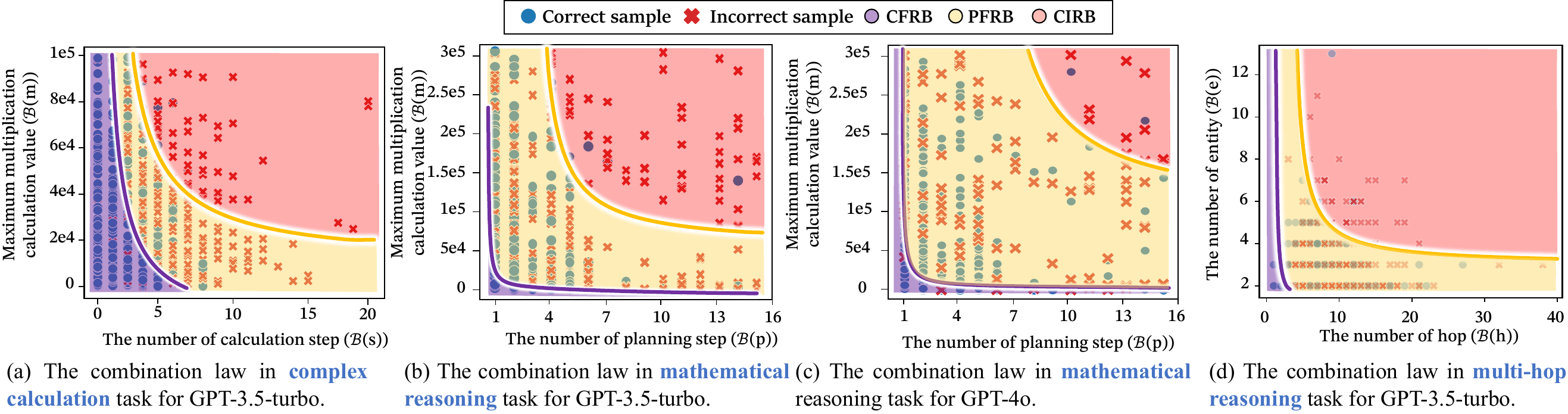}
	\caption{
		Combination law verification of RB on different tasks for RBF in textual modalities.
	}
	\label{fig:cot-ag-text}
\end{figure*}

\subsection{Categories of Reasoning Boundary}
\label{sec:types}
To guide the CoT optimization, as shown in Figure~\ref{fig:intro} (e), we categorize RBs into three categories based on empirical accuracy:
(1) \textbf{Completely Feasible Reasoning Boundary:}
A completely feasible reasoning boundary ($\CFRB{} = \mathcal{B}_{\text{Acc}\ge90\%}(t_1, t_2, \dots, t_n|m)$) is defined as the region where accuracy exceeds 90\%. This indicates that LLMs can reliably and effectively capture performance within this boundary.
(2) \textbf{Completely Infeasible Reasoning Boundary:}
In contrast, a completely infeasible reasoning boundary ($\CIRB{}=\mathcal{B}_{\text{Acc}\le 10\%}(t_1, t_2, \dots, t_n|m)$) refers to regions where accuracy falls below 10\%. Here, the model struggles to capture performance effectively.
(3) \textbf{Partially Feasible Reasoning Boundary:}
Furthermore, a partially feasible reasoning boundary ($\PFRB{}=\mathcal{B}_{10\%<\text{Acc}<90\%}(t_1, t_2, \dots, t_n|m)$) are defined as the reasoning boundary that excludes both $\CFRB$ and $\CIRB$. This boundary requires the model to engage in repeated reasoning or seek additional information to solve the problem.

	\section{Experimental Setup}
\label{sec:setting}
To evaluate the RBs of LLMs, it is vital to create a benchmark that covers a wide range of RBs. This requires tasks that are evenly distributed in complexity and reasoning steps, effectively challenging the models at their upper-bounds.
To meet this need, we introduce two novel datasets for textual modality experiments: \textsc{BigGSM} are designed with greater computational complexity and longer reasoning chains. Additionally, to address the evaluation needs of reasoning LLMs like DeepSeek-R1, we introduce an expanded benchmark, \textsc{BigGSM++}, providing a more thorough assessment of their RBs. More construction details can be found in Appendix B. For multimodal experiments, we use M3CoT~\cite{chen-etal-2024-m3cot} and OlympiadBench~\cite{he-etal-2024-olympiadbench} to evaluate across multiple domains.

Except for the expansion experiments, all experiments involving textual modality are conducted using GPT-3.5-Turbo and GPT-4o, while all multimodal experiments are performed on GPT-4o.
Aligned with Wei~\textit{et al.}~\cite{wei2022chain} and Chen~\textit{et al.}~\cite{chen-etal-2024-m3cot}, our experiment utilizes three manually constructed demonstrations for text-modal experiments and zero-shot reasoning for multi-modal experiments. Additionally, for all experiments, the top-p parameter is selected from the set ${0.95, 1}$, and the temperature parameter is chosen from the interval $[0, 1]$, with temperature serving as the primary error variable.

\section{RBF++ Analysis in Measurable Scenarios}

\subsection{Existence Verification for Reasoning Boundary}
This section examines whether an LLM exhibits varying RB across different tasks. To verify this, we evaluate the performance of LLMs on three tasks, focusing on basic arithmetic calculations, nature language planning, and code planning in mathematical tasks with textual modalities.

\subsubsection{Basic Arithmetic Calculation}
First of all, to explore the existence of RB, we analyze the RB in basic arithmetic calculation tasks, including addition, subtraction, multiplication, and division operations.
As shown in Fig.~\ref{fig:atom-rg} (a), significant performance variations emerge across three regions. For multiplication tasks, accuracy exceeds 90\% for results up to $2.2e5$, but drops below 10\% for products exceeding $2e6$. Similar RB variations are observed in other operations (Fig.1 in Appendix), confirming the presence of an RB in basic arithmetic tasks.

\subsubsection{Nature Language Planning}
Furthermore, we investigate the RB on natural language planning tasks involving mathematical reasoning. We prompt LLMs to generate detailed plans, which are then evaluated for accuracy through manual analysis. As shown in Fig.~\ref{fig:atom-rg} (b), a strong correlation exists between the number of reasoning steps and model performance. When tasks involve fewer than 2 reasoning steps, accuracy exceeds 90\%. When reasoning steps exceed 4, accuracy drops below 10\%. These results also suggest three distinct categories of RB in natural language planning tasks.

\subsubsection{Code Planning}
To further investigate RB in planning, we leverage the PAL framework~\cite{pmlr-v202-gao23f} to prompt LLMs to generate plans in code format, which are subsequently evaluated through manual annotation.
As illustrated in Fig.~\ref{fig:atom-rg} (c), code planning exhibits a similar categorization of RB as natural language planning. However, due to the clearer logical structure and reduced linguistic complexity inherent in code on textual mathematical scenarios, the accuracy of planning in this format consistently exceeds that of natural language planning tasks.

\begin{figure*}[t]
	\centering
	\includegraphics[width=0.99\textwidth]{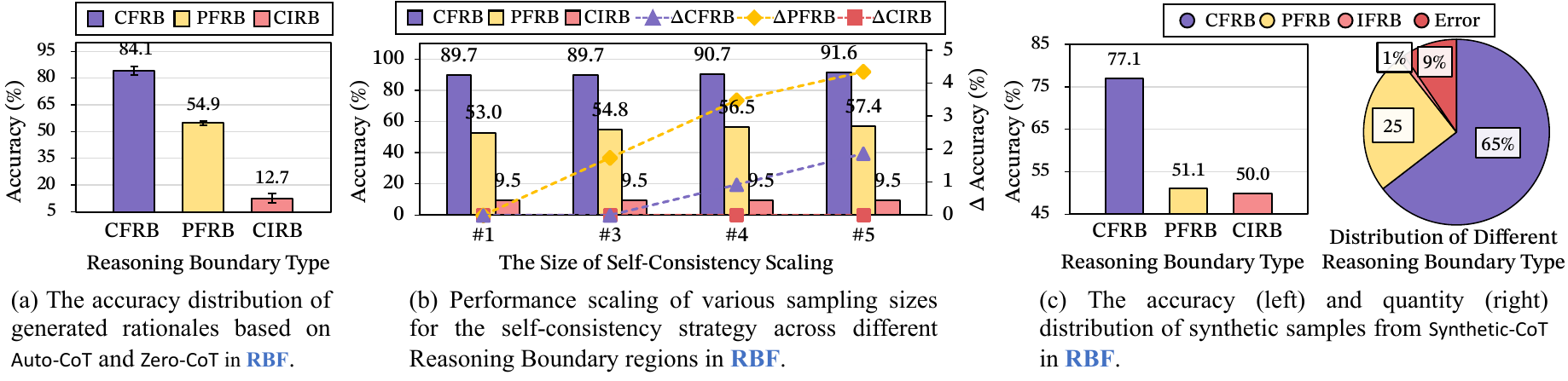}
	\caption{
		Nature analysis at different reasoning boundaries on \textsc{BigGSM} with text-modal scenarios. For Fig. (c), all samples in CIRB are special value points, like $25000 \times 1000$. In fact, no real CIRB samples are obtained.
	}
	\label{fig:rb-nature-text}
\end{figure*}

\subsection{Combination Law Verification on Different Tasks}

\subsubsection{Combination Law in Complex Arithmetic Calculation}
Building on Eq.~\eqref{eq:combine-law-simple}, we hypothesize that the combination law for RB in complex arithmetic calculations is the harmonic average of the arithmetic calculation RB and the calculation planning RB. To test this, we design an experiment focused on formulas involving addition, subtraction, and multiplication, such as ``$(1+2)*3$''.
Since the complexities of addition and subtraction are estimated to be around $1e{15}$ (as shown in Fig. 1 in Appendix), the arithmetic calculation RB is primarily influenced by the multiplication RB and the calculation planning RB. As shown in Fig.~\ref{fig:cot-ag-text} (a), two distinct RB lines, $\mathcal{B}_{Acc=90\%}$ and $\mathcal{B}_{Acc=10\%}$, align with the combination law of these basic RBs, as defined by Eq.~\eqref{eq:combine-law}. Additionally, these lines clearly categorize the RBs into three distinct areas.

\subsubsection{Combination Law in Mathematical Reasoning}
Practically, we design an experiment focused on real mathematical reasoning within the annotated \textsc{BigGSM} benchmark, expanding the range of arithmetic and step numbers. Building on prior work~\cite{tan2023causal,xiao2024theory}, we propose that natural language mathematical reasoning involves two key sub-tasks: step planning and step calculation, which correspond to global logical planning and local mathematical computation, respectively. Each output step in the model corresponds to a basic operation, with step calculations limited by the maximum number of multiplications ($\mathcal{B}(m)$), where $\mathcal{B}(c) \approx \mathcal{B}(m)$. Formally, let the step planning RB be denoted as $\mathcal{B}(p)$ and the step calculation RB as $\mathcal{B}(c)$. The combined RB is given by:
\begin{equation}
	\mathcal{B}^{\texttt{CoT}}(c, p) = \frac{1}{\frac{1}{\mathcal{B}(c)} + \frac{1}{\mathcal{B}(p)}}.
	\label{eq:cot}
\end{equation}
As depicted in Fig.~\ref{fig:cot-ag-text} (b,c), the performance distribution of the RB (comprising $\mathcal{B}_{Acc=90\%}$ and $\mathcal{B}_{Acc=10\%}$) fully matches the theoretical combination law in Eq.~\eqref{eq:cot}.

\subsubsection{Combination Law in Multi-hop Reasoning}
Beyond mathematics, we extend the combination law beyond mathematics to multi-hop question answering. Specifically, we apply this combination law to the HotpotQA~\cite{yang2018hotpotqa}, where we define the unified RB as a combination of global hop-planning RB and local knowledge entity RB. As shown in Fig.~\ref{fig:cot-ag-text} (d), $\mathcal{B}_{Acc=90\%}$ and $\mathcal{B}_{Acc=10\%}$ adhere to the weighted harmonic mean of these two sub-RBs. This demonstrates that, in addition to math-related tasks, multi-hop question answering also satisfies the proposed combination law, revealing three distinct RBs.

\subsection{Nature Analysis for Different Reasoning Boundaries}
Based on the definitions of various RBs, we divide the area into three parts. This section examines whether the defined RBs align with the inherent characteristics of the model. We analyze the nature of each RB in detail:

\subsubsection{\CFRB{} means complete mastery even without demonstration.}
According to its definition, a question within \CFRB{} indicates a deep understanding of the issue by the LLM. To verify this, following \cite{zhang2022automatic} and \cite{wei2022chain}, we formulate a mathematical request and generate chain-of-thought reasoning and answers through zero-shot prompting, without any demonstration. As shown in Fig.~\ref{fig:rb-nature-text} (a), \CFRB{} achieves a 29.2\% improvement in generating correct rationale compared to other RBs, confirming that the model can effectively master tasks within this boundary.

\subsubsection{\PFRB{} means moderate confidence in its solution and needs consensus building process.}
\PFRB{} reflects a moderate confidence level, requiring a consensus-building process. This is similar to human decision-making, where moderate confidence often necessitates multiple consensus. Inspired by this, we apply the Self-Consistency (\texttt{SC}) approach~\cite{wang2022self}, which combines multiple reasoning paths to reach a final answer. As shown in Fig.~\ref{fig:rb-nature-text} (b), the accuracy improves significantly as the number of reasoning paths increases within \PFRB{}, suggesting that the model exhibits moderate confidence and needs multiple voting.

\subsubsection{\CIRB{} exhibits poor performance even with consensus building.}
As illustrated in Fig.~\ref{fig:rb-nature-text} (a), questions in \CIRB{} show extremely low accuracy (around 9.5\%). The model's performance remains consistently poor, with no improvement in \texttt{SC} for this boundary, as seen in Fig.~\ref{fig:rb-nature-text}. This indicates that \CIRB{} represents a boundary where the model struggles to reason effectively, even with consensus building.

\begin{figure*}[t]
	\centering
	\includegraphics[width=0.99\textwidth]{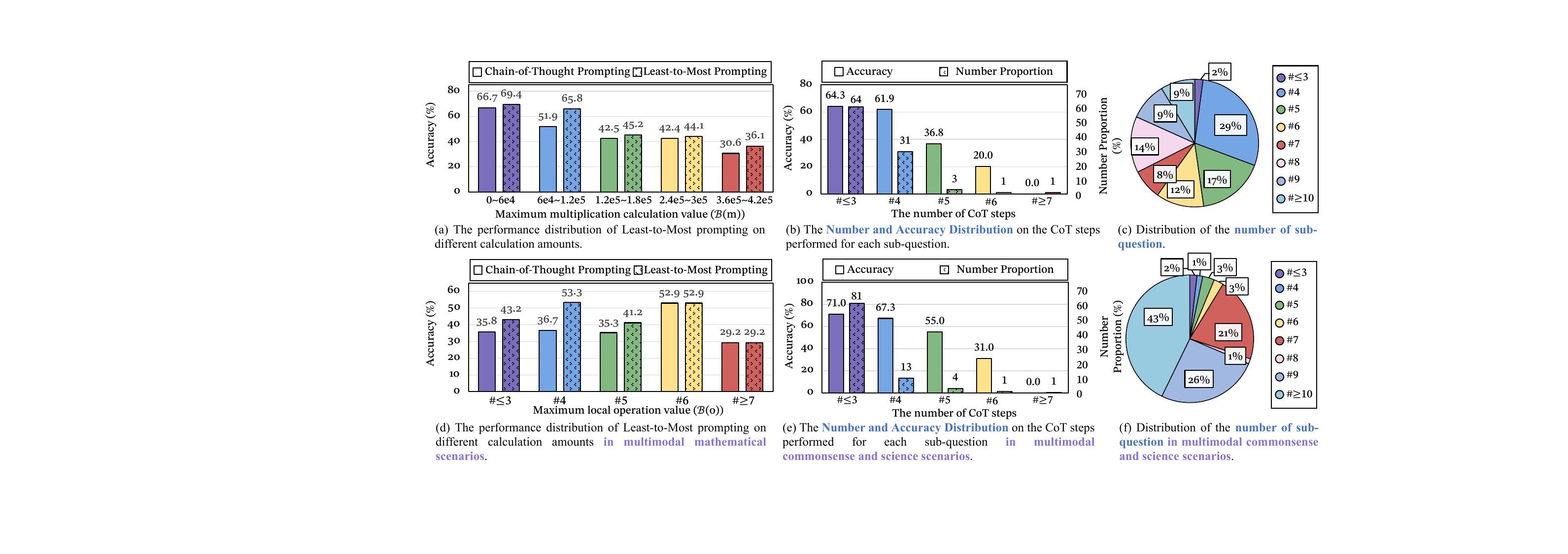}
	
	\captionof{figure}{The performance analysis of Least-to-Most prompting in textual modalities.}
	\label{fig:least-to-most-text}
\end{figure*}

\begin{figure}
	\centering
	\includegraphics[width=0.96\linewidth]{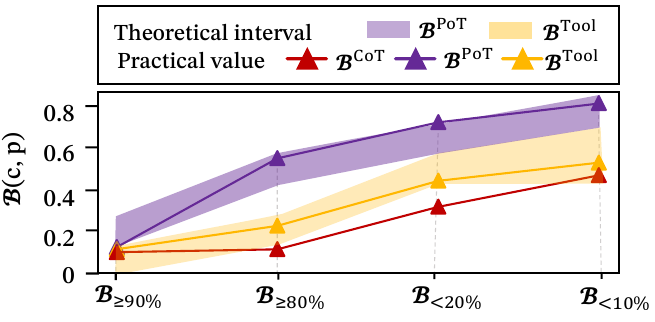} 
	\captionof{figure}{Analysis of the impact of \texttt{CoT}, \texttt{TU}, and \texttt{PoT} on reasoning boundary $\mathcal{B}(c, p)$ in textual scenarios.
	}
	\label{fig:upperbound-text}
\end{figure}

\subsubsection{LLM has self-awareness of its own RBs.}
In parallel, a natural question arises: \textit{Can the model recognize its inherent RBs?} To explore this, we use the \texttt{Synthetic-CoT} method~\cite{shao2023synthetic} to prompt the LLM to generate CoT data. As shown in Fig.~\ref{fig:rb-nature-text} (c), over 65\% of the generated samples fall within \CFRB{}, showing a clear performance advantage over other RBs. This suggests that LLMs possess an inherent understanding of their RBs, which enables them to identify the tasks they can potentially self-assess their capabilities.

\subsection{How can we improve CoT by optimizing RB?}
In our framework, RB plays a vital role in reflecting model's performance. Enhancing CoT can be effectively achieved by optimizing the RB calculation, denoted as $\mathcal{B}(c)$. This section adopts this perspective to explain why Program-of-Thought outperforms Tool-Usage in textual modalities~\cite{yao2023react, chen2023program}, and demonstrates how RB optimization can further improve CoT.

\subsubsection{Tool Usage (\texttt{TU}) is an effective method to boost RB value for an LLM in textual modalities.}
When a model employs external tools~\cite{paranjape2023art}, it is reasonable to assume that calculations can be performed with infinite precision. This assumption causes the RB related to mathematical operations to approach infinity, i.e., $\mathcal{B}(c) \rightarrow +\infty$. Thus, the overall RB can be expressed as:
\begin{equation}
	\mathcal{B}^{\texttt{T}}(c, p)\! = \!\!\lim\limits_{\mathcal{B}(c) \rightarrow +\infty}\frac{1}{\frac{1}{\mathcal{B}(c)}\! +\! \frac{1}{\mathcal{B}(p)}} \!= \mathcal{B}(p). \label{eq:tool-usage}
\end{equation}
It is obvious that, $\mathcal{B}^{\texttt{T}}(c, p) > \mathcal{B}^{\texttt{CoT}}(c, p)$, indicating that \texttt{TU} enhances the RB value. This explains why \texttt{TU} outperforms vanilla CoT, as shown in Tab.~\ref{exp:main-exp}.
Additionally, as illustrated in Fig.~\ref{fig:upperbound-text} (a), the distribution of the theoretical RB closely matches the actual observed RB, further confirming the reliability of our theoretical framework.

\begin{figure}[t]
	\centering
	\includegraphics[width=0.99\linewidth]{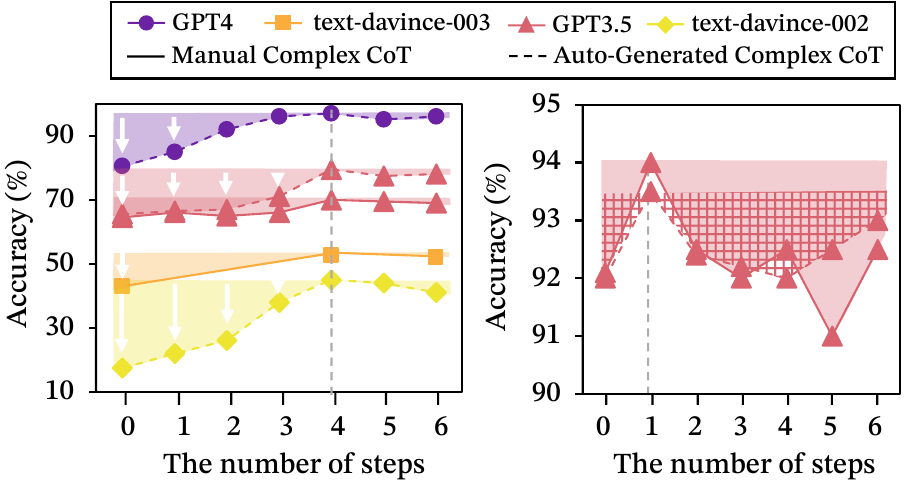}
	
	\captionof{figure}{Correlation between the number of steps and performance of \texttt{CCoT} on GSM8K (left) and SingleEq (right) from Jin~\textit{et al.}~\cite{jin2024impact} and Fu~\textit{et al.}~\cite{fu2023complexitybased}.
	}
	\label{fig:complex-cot-text}
\end{figure}

\subsubsection{Program-of-Thought (\texttt{PoT}) further enhances RB value in textual modalities}
Eq.~\eqref{eq:tool-usage} shows that the RB of an LLM with tools in calculation tasks depends on its planning ability. However, the verbosity of natural language can hinder this planning, as noted in previous studies~\cite{pmlr-v202-gao23f, hu2023code}. The \texttt{PoT} framework~\cite{chen2023program} addresses this by using code to represent logic more concisely, enabling better planning (see Fig.~\ref{fig:atom-rg} (b, c)).
This shift results in more refined planning reasoning, denoted as $\mathcal{B}^{*}(p) > \mathcal{B}(p)$. As a result, the PoT-based reasoning boundary, $\mathcal{B}^{\texttt{PoT}}(c,p)$, outperforms \texttt{TU}, $\mathcal{B}^{\texttt{Tool}}(c,p)$, as seen in the performance improvements with \texttt{PoT} (Tab.~\ref{exp:main-exp}). Moreover, Fig.~\ref{fig:upperbound-text} confirms that both the theoretical and practical reasoning boundaries of PoT exceed those of \texttt{TU}, highlighting PoT's theoretical advantages and empirical effectiveness.

\subsection{How can we improve CoT based on a certain RB?}

Optimizing the RB is crucial for enhancing CoT reasoning. However, effective CoT optimization requires modifications to the model or its underlying reasoning structure. Thus, it is essential to adjust the reasoning path to match the optimized RB, denoted as $d^{*}=\mathcal{B}_{Acc=K_1}$, where $K_2<K_1$ represents the original, less optimized RB, $\mathcal{B}$. According to Eq.~\eqref{eq:cot}, $\mathcal{B}$ is influenced by both arithmetic and planning-related RBs. We explore two main strategies for improving reasoning abilities, as detailed below (for further analysis, see Appendix~C):

\subsubsection{Complex CoT (\texttt{CCoT})}
The \texttt{CCoT} approach aims to reduce the arithmetic RB by decomposing large steps into smaller, simpler ones. This strategy also increases the planning boundary. By breaking tasks into more manageable steps, \texttt{CCoT} eases the load on individual calculations, lowering the overall value of $d$. However, this method introduces a trade-off: while it reduces the pressure on individual steps, it increases the number of planning steps, thus adding pressure during the planning phase. As shown in Fig.~\ref{fig:complex-cot-text}, the performance of LLMs initially improves but deteriorates as the number of \texttt{CCoT} demonstration steps grows, each with varying levels of reasoning complexity, particularly in multi-modal scenarios. To Achieve optimal performance with \texttt{CCoT}, it requires a careful balance between the number of reasoning steps and the associated planning pressure.

\begin{table}[t]
	\centering
	\captionof{table}{Main experimental results on GPT-3.5-Turbo.}
		\begin{adjustbox}{width=0.99\linewidth}
			\begin{tabular}{lccc}
				\toprule
				\multirow{2}{*}{Model}  & \multicolumn{3}{c}{\textsc{BigGSM}} 
				\\\cmidrule{2-4}
				& Acc. ($\uparrow$) & Input Token ($\downarrow$) & Output Token ($\downarrow$) \\
				\midrule
				\texttt{CoT} & 57.00  & 780.43 & 96.76  \\
				\midrule
				\rowcolor{gray!8}\multicolumn{4}{c}{RB-Optimized Methods}\\
				\midrule
				\texttt{Tool Usage} & 71.64  & 688.43 & 129.53  \\
				\texttt{PoT} & 78.25  & 657.43 & 78.25
				  \\
				\midrule
				\rowcolor{gray!8}\multicolumn{4}{c}{Reasoning-Path-Optimized Methods}\\
				\midrule
				\texttt{Least-to-most}  & 58.25  & 679.59 & 176.09 
				\\
				\texttt{Complex-CoT}  & 59.78  & 1111.43 & 131.82 
				\\
				\midrule
				\texttt{CoT+MARP}& 64.37  & 614.43 & 95.12 
				\\
				\texttt{PoT+MARP} & \textbf{80.55}  & \textbf{576.43} & \textbf{76.34} 
				\\
				
				\bottomrule
			\end{tabular}
		\end{adjustbox}

	\label{exp:main-exp}
\end{table}

\begin{figure}[t]
	\centering
	\includegraphics[width=0.99\linewidth]{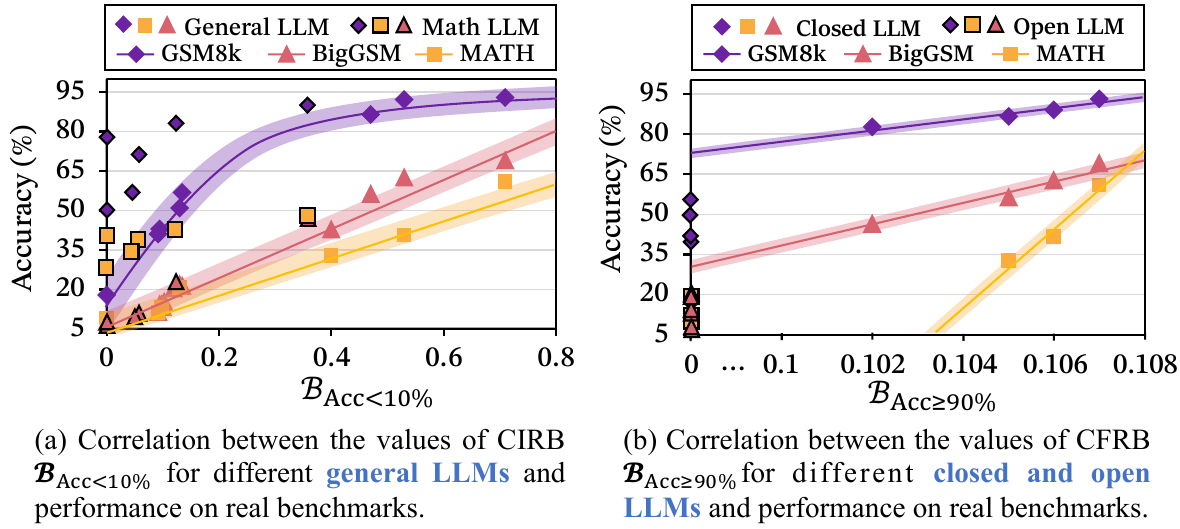}
	\captionof{figure}{Correlation between the values of RB for different models and performance on real benchmarks.}
	\label{fig:correlation-text}
\end{figure}

\subsubsection{Least-to-Most (\texttt{LtM})}
\texttt{LtM} approach reduces the pressure of local planning by breaking the problem into smaller sub-questions, effectively decreasing the planning boundary and reducing the value of $d$. It improves local operation accuracy in both text-based and multi-modal scenarios, as shown in Fig.~\ref{fig:least-to-most-text} (a). Specifically, by minimizing local operation steps, as shown in Fig.~\ref{fig:least-to-most-text} (b), LLMs exhibit better performance and are more likely to generate simpler sub-questions, which further simplify local operations. However, while \texttt{LtM} reduces local planning pressure, it increases global planning pressure due to the larger number of generated sub-questions, as depicted in Fig.~\ref{fig:least-to-most-text} (c). Therefore, it indicates the fact that, while local planning pressure is alleviated, \texttt{LtM} always does not effectively mitigate global planning challenges with limited performance.

\subsection{Minimum Acceptable Reasoning Paths (MARP)}
To address the limitations discussed earlier, we propose the Minimum Acceptable Reasoning Paths (MARP) prompting method. The optimization goal is to reduce relevant computational and planning burdens by framing problems within specific RBs, thus improving performance. Additionally, MARP aims to increase model acceptability by optimizing the computation per demonstration step and reducing the number of global planning steps, thereby alleviating planning pressure. As shown in Tab.~\ref{exp:main-exp}, MARP enhances model performance while reducing token consumption. By maximizing operations per step, MARP simplifies the problem-solving process and improves efficiency. Detailed prompts are provided in Appendix Appendix~D.

\subsection{Reasoning Boundary Scaling for Textual LLM}
To broaden the applicability of our mechanism, we test it on 27 models (details in Table~I in Appendix). As shown in Fig.~\ref{fig:correlation-text} (a), a positive correlation is observed between models' RB and corresponding accuracy on mathematical benchmarks. Additionally, models trained with mathematical data, such as MathInstruct for SFT, present notable outliers distinct from typical LLMs. Despite this, they maintain a strong positive correlation with reasoning boundaries, indicating that mathematically focused training influences model behavior.
However, Fig.~\ref{fig:correlation-text} (b) reveals further insights. Notably, the primary difference between open-source and closed-source models lies in the \CFRB, with all models, except the closed-source one, showing a \CFRB{} of 0. These points are quite potential areas for optimization.

\section{Reasoning Model Exploration}

\begin{figure}[t]
	\centering
	\includegraphics[width=0.99\linewidth]{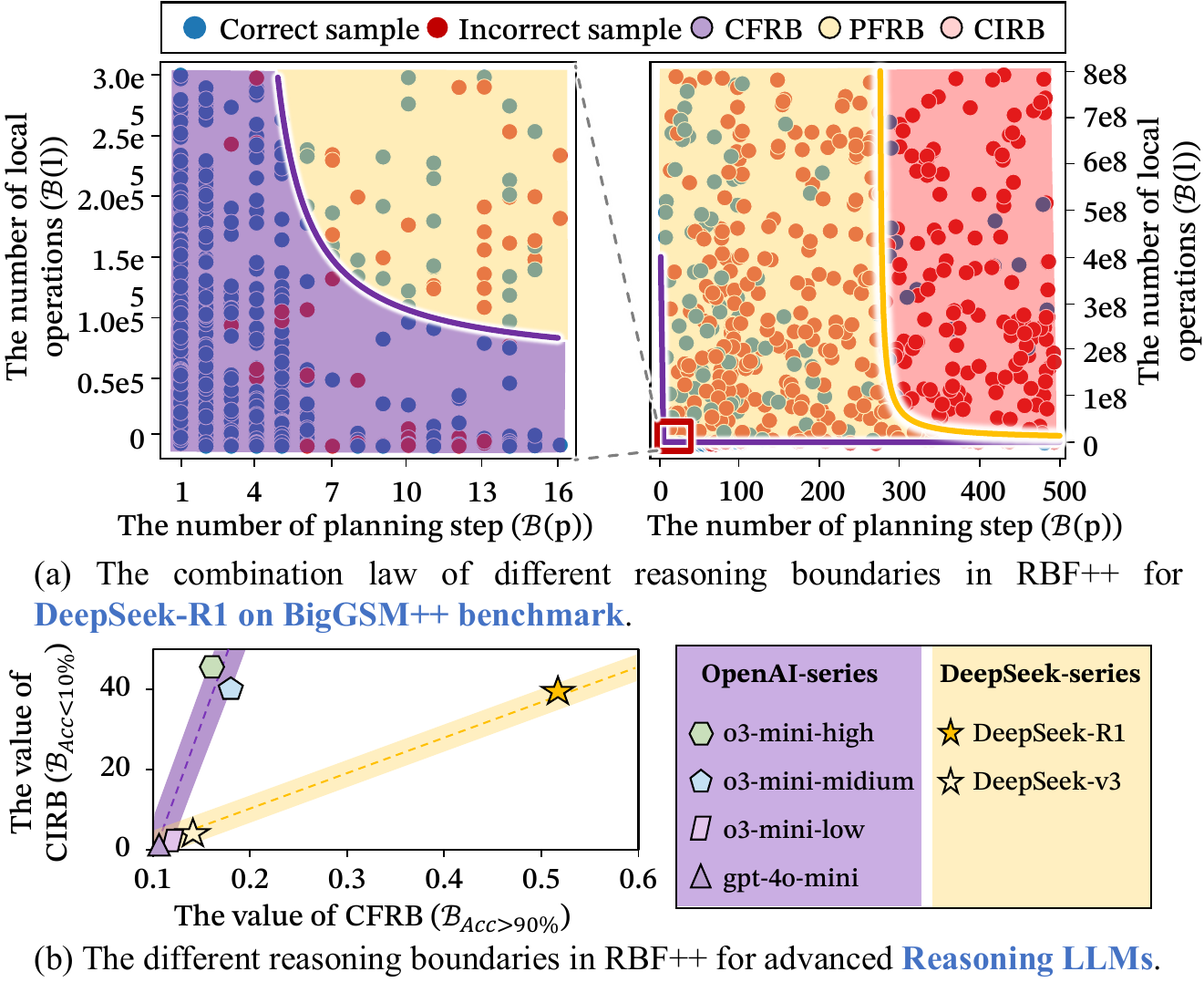}
	\caption{
		Correlation between the values of RB for different models and performance on real benchmarks.
	}
	\label{fig:reasoning-cot-rg}
\end{figure}

To better understand advanced reasoning LLMs~\cite{chen2025towards}, we introduce the \textsc{BigGSM++} to examine the RBs of reasoning LLMs like DeepSeek-R1 and OpenAI-o1 via \texttt{RBF++}. As shown in Fig.~\ref{fig:reasoning-cot-rg} (a), the \CFRB{} significantly enhances performance, with a minor improvement compared to the \CIRB{} in \textsc{BigGSM++}, which is nearly 100 times more efficient than DeepSeek-v3, its base model. This improvement is largely attributed to the effectiveness of advanced Reinforcement Learning and Inference Scaling strategies, which have a more profound impact on \CFRB{} than typical enhancements, suggesting avenues for further research.

\begin{figure*}[t]
	\centering
	\includegraphics[width=0.99\textwidth]{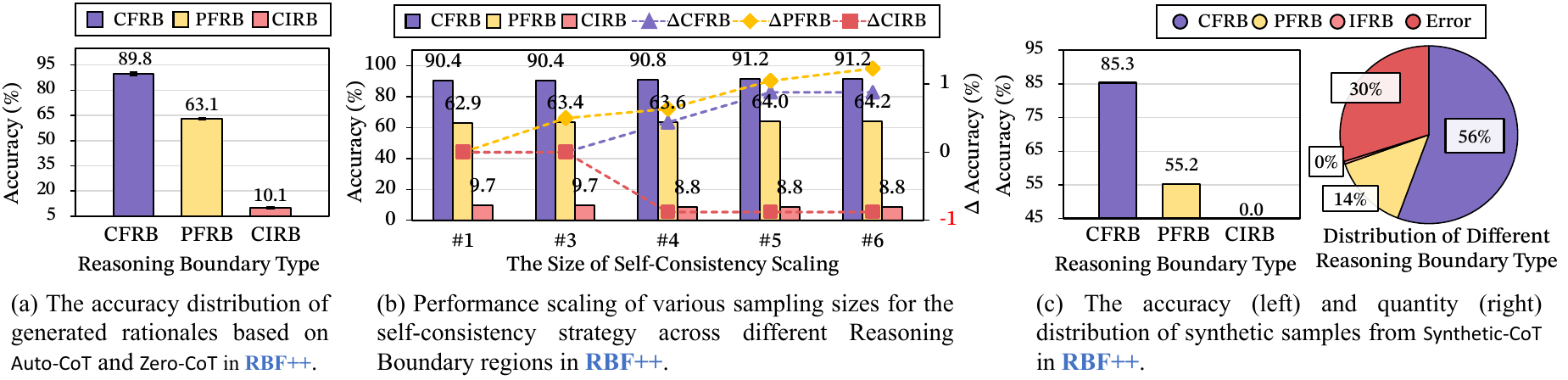}
	\caption{
		Nature analysis at different reasoning boundaries for unmeasurable boundaries on M3CoT with multi-modal scenarios.
	}
	\label{fig:nature-multimodal}
\end{figure*}

Additionally, as shown in Fig.~\ref{fig:reasoning-cot-rg} (b), a notable observation arises: the multi-stage supervised-finetuning and RL approach in DeepSeek-R1 achieves a more balanced improvement in both the \CFRB{} and \CIRB{}. In contrast, o3-mini primarily enhances the \CIRB{}, enabling the model to solve more previously unsolved problems. However, this improvement does not lead to mastery in the area, meaning the model continues to struggle with mastering medium-level challenges that basic LLMs can partially address. As a result, o3-mini underperforms relative to DeepSeek-R1 in medium-complexity tasks like MATH, which focus more on the \CFRB{}. However, o3-mini outperforms or matches DeepSeek-R1 in more complex tasks such as AIME, where the focus is on the \CIRB{}~\cite{jahin2025unveiling}.

\section{RBF++ Analysis in Unmeasurable Scenarios}
\subsection{Constant Assumptions on Multimodal Scenarios}
For a vertical domain problem based on Multimodal CoT reasoning~\cite{duan2023deeplogic,hong2023visual}, the process can be divided into four primary reasoning boundaries: task step planning RB, $\mathcal{B}(p)$, operation RB, $\mathcal{B}(o)$ for each step, vertical domain RB, $\mathcal{B}(v)$. These unmeasurable boundaries can be combined as follows:
\begin{equation}
	\mathcal{B}(o,p,v) = \frac{1}{\frac{1}{\mathcal{B}(p)}+\frac{1}{\mathcal{B}(o)}+\frac{1}{\mathcal{B}(v)}},
\end{equation}
where $\mathcal{B}(v)$ are difficult to measure directly.
To tackle this, we propose an alternative method to measure reasoning boundaries, allowing the model to generate direct answers without relying on CoT reasoning steps. In this approach, reasoning depends only on unmeasurable vertical domain RB $\mathcal{B}(v)$, not on task step planning RB $\mathcal{B}(p)$ or operation RB $\mathcal{B}(o)$.
We evaluate this by assessing the model's performance using direct prompts without CoT outputs. The resulting performance is used to measure $\mathcal{B}(v)$ as ${z_1}$, which aids in determining the corresponding constant.
Thus, the combined RB is defined as:
\begin{equation}
	\mathcal{B}(p,o,v) = \frac{1}{\frac{1}{\mathcal{B}(p)}+\frac{1}{\mathcal{B}(o)}+{z_1}}.
	\label{eq:constant}
\end{equation}

\begin{figure}[t]
	\centering
	\includegraphics[width=0.99\linewidth]{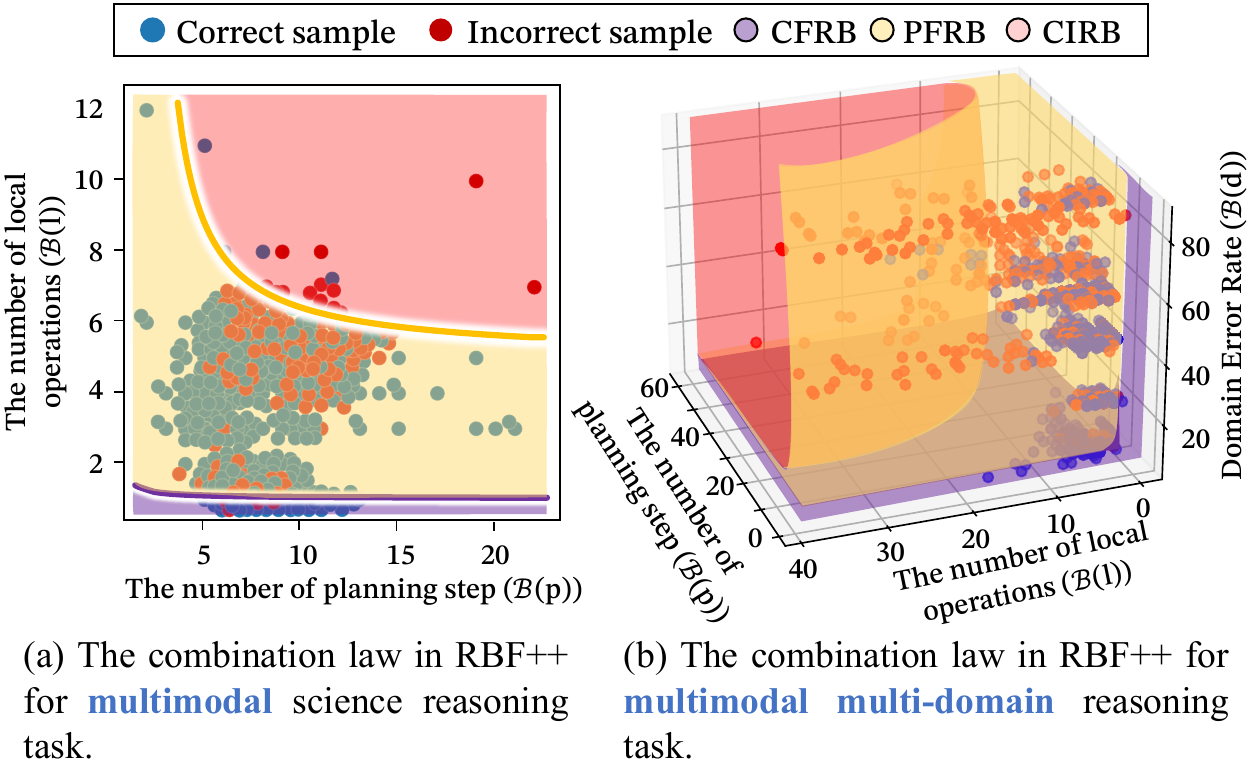}
	\caption{
		Combination law verification of RB on different tasks across multiple modalities.
	}
	\label{fig:cot-rg-multimodal}
\end{figure}

As shown in Fig.~\ref{fig:cot-rg-multimodal} (a), we validate our results within the nature science topic of M3CoT~\cite{chen-etal-2024-m3cot}.
Two distinct RBs, namely $\mathcal{B}_{Acc=90\%}$ and $\mathcal{B}_{Acc=10\%}$, clearly emerge, both of which satisfy the conditions in Eq.~\eqref{eq:constant}. These results align with the combination law for RBs, supporting the constant assumption in RBF++, and enabling the application of adaptive unmeasurable scenarios in multimodal settings.

\subsection{Reasoning Boundary Division on Multimodal Scenarios}
Based on the previous analysis, we divide the unmeasurable RB within multiple modalities and domains RB, $\mathcal{B}_v$, into two components: the domain knowledge RB, $\mathcal{B}_{k}$, and the multimodal perception RB, $\mathcal{B}_{mm}$, satisfying:
\begin{equation}
	\mathcal{B}(v) = B(k,mm) = \frac{1}{\frac{1}{\mathcal{B}_{k}} + \frac{1}{\mathcal{B}_{mm}}}.
\end{equation}
Given the challenges in quantifying the multimodal perception RB, we assume it as a constant. When multimodal perception requirements are stable across the dataset, we set this RB as a constant $z$, resulting in:
$
	\mathcal{B}(v) = \frac{1}{\frac{1}{\mathcal{B}(k)} + z'},
$
where$z'=\frac{1}{z}$. Consequently, the combined RB can be extended as follows:
\begin{equation}
	\mathcal{B}(o,p,v) = \frac{1}{\frac{1}{\mathcal{B}(p)}+\frac{1}{\mathcal{B}_o}+\frac{1}{\mathcal{B}(k)} + z'}.
	\label{eq:vertical-domain-rb}
\end{equation}

Our experimental results, shown in Fig.~\ref{fig:cot-rg-multimodal} (b), reveal two distinct RBs, clearly observed in the $\mathcal{B}_{Acc=90\%}$ and $\mathcal{B}_{Acc=10\%}$. These distributions align with the theoretical constraints in Eq.~\eqref{eq:vertical-domain-rb}, supporting the reasoning boundary combination law. This consistency validates the effectiveness of the RBF++ framework's division mechanism, ensuring its applicability across multimodal environments and heterogeneous domains within a unified theoretical framework.

\begin{figure*}[t]
	\centering
	\includegraphics[width=0.99\textwidth]{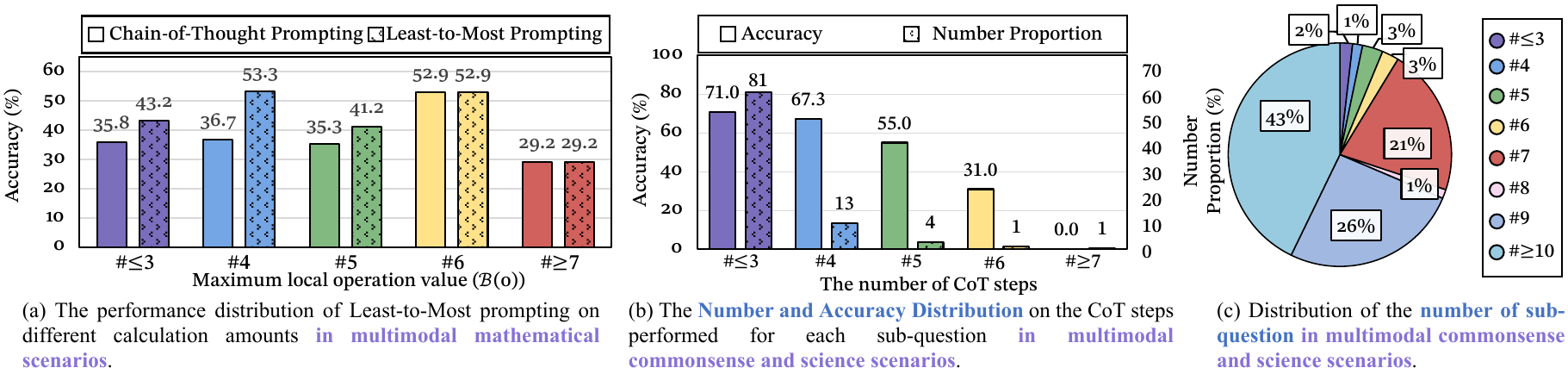}
	
	\captionof{figure}{The performance analysis of Least-to-Most prompting in multimodal scenarios.}
	\label{fig:least-to-most-multimodal}
\end{figure*}

\begin{figure}
	\centering
	\includegraphics[width=0.99\linewidth]{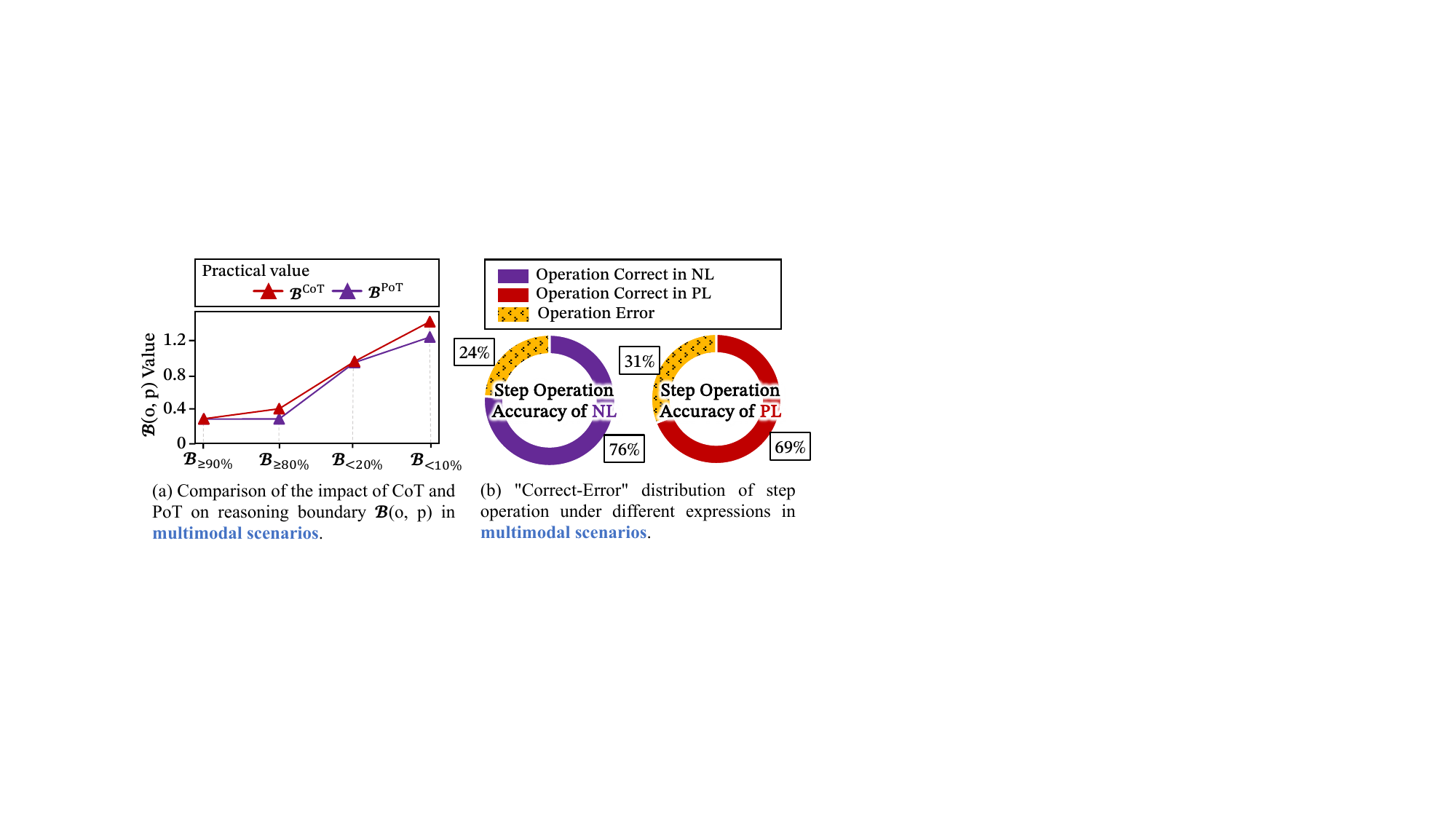} 
	\captionof{figure}{Analysis of the impact of PoT in multimodal scenarios.
	}
	\label{fig:upperbound-multimodal}
\end{figure}

\subsection{Similar Nature for Different Reasoning Boundaries}
\label{sec:nature}
In this section, we delve into an analysis of the intrinsic nature of the three distinct RBs for unmeasurable scenarios.

\subsubsection{\CFRB{} also implies mastery even without the need for demonstrations in multimodal scenarios} As previously discussed, questions within the \CFRB{} category require a comprehensive understanding of the task, in textual modality. Extending this to the multimodal context, as shown in Fig.~\ref{fig:nature-multimodal} (a), the model exhibits at least a 26.7\% improvement in performance for \CFRB{} questions within  multimodal scenarios.

\subsubsection{\PFRB{} also reflects moderate confidence, requiring consensus-building in multimodal scenarios}
Similarly, as illustrated in Fig.~\ref{fig:nature-multimodal} (a), model performance in \PFRB{} remains moderate. Moreover, applying \texttt{SC} notably improves accuracy for \PFRB{} in multimodal contexts, as shown in Fig.~\ref{fig:nature-multimodal} (b), further confirming this property in multimodal scenarios.

\subsubsection{\CIRB{} also exhibits extremely poor reasoning performance despite consensus-building in multimodal scenarios}
As depicted in Fig.~\ref{fig:nature-multimodal} (a), the accuracy in \CIRB{} is approximately 10\%, and this does not improve and even experiences a decline in performance in multimodal scenarios (Fig.~\ref{fig:nature-multimodal} (b)). These results emphasize that the poor reasoning performance of \CIRB{} persists across modalities, where the model’s reasoning remains inadequate, regardless of consensus-building.

\subsubsection{LLM also has self-awareness of its own RBs}
In multimodal contexts, as shown in Fig.~\ref{fig:nature-multimodal} (c), over 56\% of the samples fall under \CFRB{}, achieving 85.3\% accuracy. Less than 1\% of samples belong to \CIRB{}, with the lowest performance. These results confirm that the model's self-awareness of its RBs and its ability to assess them remain consistent across both textual and multimodal contexts.

\subsection{Why are these effective prompting in textual scenarios not effective in Multimodal scenarios?}
\subsubsection{Program-of-Thought fails on multimodal scenarios due to its inability to optimize local operation RB}
While the \texttt{PoT} framework effectively improves reasoning in textual tasks, its capability in multimodal scenarios remains limited. This limitation stems from \texttt{PoT}'s reliance on code-based logic representation, which excels in sequential and symbolic reasoning but struggles with the parallel, spatial, and perceptual reasoning necessary for visual modalities. Consequently, although \texttt{PoT} enhances global RBs in text scenarios, it fails to optimize local operation RB, where fine-grained, modality-specific processing is essential.
Empirical results in Fig.~\ref{fig:upperbound-multimodal} (a) confirm this issue: the reasoning gains observed in textual tasks do not transfer to multimodal tasks, with performance even falling below that of vanilla natural language approaches. Further analysis in Fig.\ref{fig:upperbound-multimodal} (b) reveals a 7\% increase in step operation errors, suggesting that the structured, code-driven reasoning of \texttt{PoT} is poorly aligned with the unstructured nature of visual data.

\begin{figure}[t]
	\centering
	\includegraphics[width=0.99\linewidth]{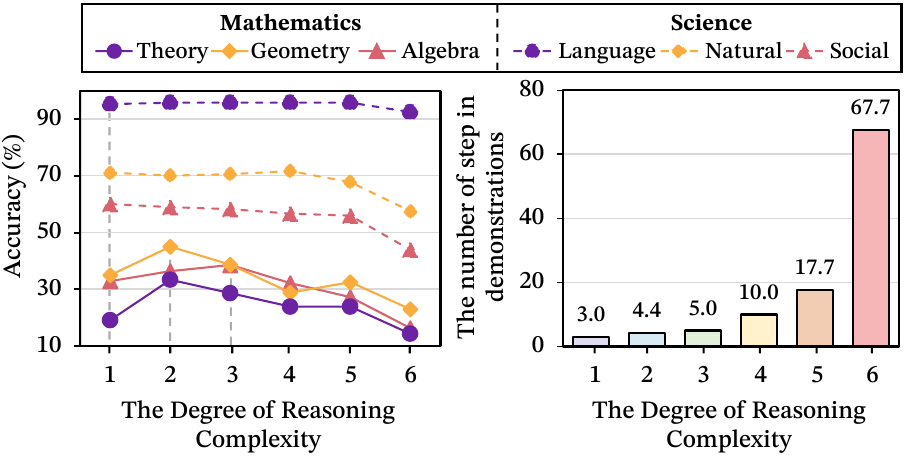}
	
	\captionof{figure}{Correlation between the number of steps and performance of Complex-CoT on M3CoT.}
	\label{fig:complex-cot-1}
\end{figure}

\begin{table*}[t]
	\centering
	\caption{Main experimental results on GPT-4o. See Appendix~D for more prompting details.}
	\begin{adjustbox}{width=0.99\textwidth}
	\begin{tabular}{lcccccccccc}
		\toprule
		\multirow{2}{*}{Prompt}  & \multicolumn{3}{c}{\textsc{Science}} & \multicolumn{3}{c}{\textsc{Commonsense}} & \multicolumn{3}{c}{\textsc{Mathematics}} & \multirow{2}{*}{Total}  
		\\\cmidrule{2-10}
		& Lang & Natural & Social & Physical & Social & Temporal & Algebra & Geometry & Theory & \\
		\midrule
		\texttt{CoT}~\cite{kojima2022large} & 94.79 & 74.07 & 56.85 & 90.00 & 76.03 & 86.99 & 39.29 & 28.75 & 23.81 & 68.68 \\
		\texttt{Least-to-Most}~\cite{zhou2022least} & 88.63 & 66.67 & 56.05 & 91.11 & 76.86 & 83.74 & 37.14 & \textbf{45.00} & 33.33 & 65.88
		\\
		\texttt{Complex-CoT}~\cite{fu2023complexitybased}  & 95.73 & 71.65 & 56.69 & 85.56 & 73.14 & 86.18 & 32.14 & 28.75 & 23.81 & 66.95
		\\
		\midrule
		\rowcolor{gray!8}\texttt{MARP}~\cite{chen2024unlocking} & 96.21 & 68.84 & 58.44 & 86.67 & 76.86 & 86.99 & 39.29 & 37.50 & 33.33 & 67.82
		\\
		\rowcolor{gray!8}\texttt{MARP++} & \textbf{96.21} & \textbf{78.93} & \textbf{65.13} & \textbf{92.22} & \textbf{77.69} & \textbf{87.80} & \textbf{43.57} & 41.25 & \textbf{33.33} & \textbf{73.77}
		\\
		
		\bottomrule
	\end{tabular}
\end{adjustbox}

\label{exp:main-exp-2}
\end{table*}

\subsubsection{Complex CoT (\texttt{CCoT}) fails due to over-complexity}
\texttt{CCoT} reduces arithmetic RB and local operation RB by breaking complex steps into simpler sub-steps, thereby extending the planning boundary. However, excessive fragmentation increases planning pressure in multimodal scenarios. As illustrated in Fig.~\ref{fig:complex-cot-1}, performance peaks at an intermediate step count in multimodal scenarios before declining due to overly complex demonstrations. The optimal \texttt{CCoT} thus requires a balance between reasoning granularity and planning strain. As shown in Tab.~\ref{exp:main-exp-2}, \texttt{CCoT} succeeds in complex mathematical tasks but fails on simpler ones.

\subsubsection{Least-to-Most (\texttt{LtM}) fails due to excessive sub-question division}
\texttt{LtM} reduces local planning pressure by breaking down problems into simpler sub-questions, thereby narrowing the planning scope. However, as shown in Figs.~\ref{fig:least-to-most-multimodal} (a, b), this enhances local accuracy in both textual and multimodal contexts. In multimodal scenarios, Fig.~\ref{fig:least-to-most-multimodal} (c) illustrates that \texttt{LtM} extremely increases global planning pressure by generating numerous sub-questions (over 43\% with more than 10 sub-questions), failing by surpassing global optimization RBs.

\begin{figure}[t]
	\centering
	\includegraphics[width=0.99\linewidth]{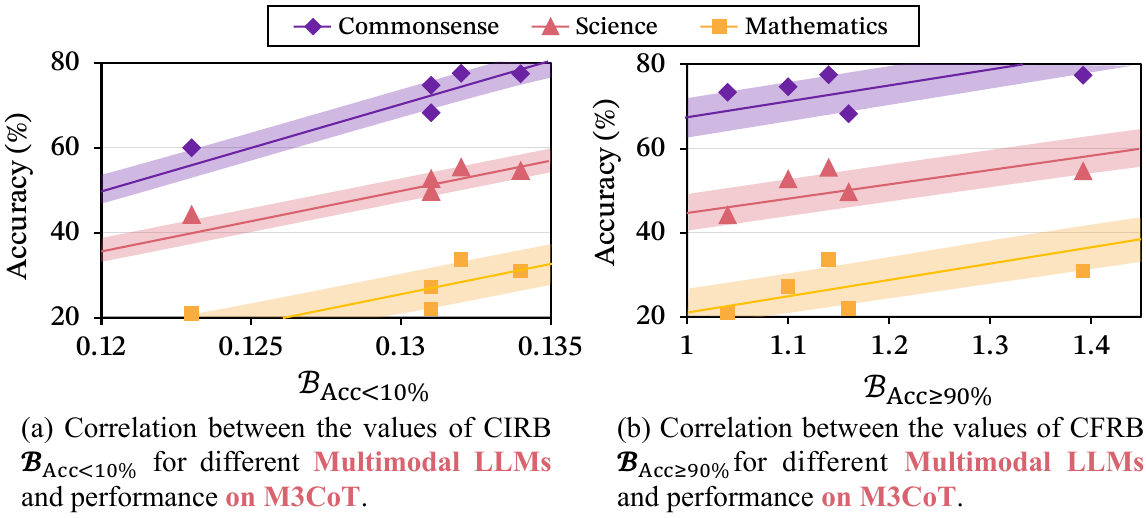}
	\captionof{figure}{Correlation between RB values for different models and performance on M3CoT.}
	\label{fig:correlation-multimodal}
\end{figure}

\subsection{Minimum Acceptable Reasoning Paths++ (MARP++)}
In multimodal scenarios, we are required to optimize the CoT for unmeasurable capabilities, like commonsense, scientific knowledge and multimodal perception, which are ignored by \texttt{MARP}. As shown in Tab.~\ref{exp:main-exp-2}, \texttt{MARP} improves performance primarily in mathematical domains, while its results in commonsense and science remain suboptimal. Furthermore, to optimize the critical but unmeasurable aspects such as visual perception and knowledge boundaries, we introduce MARP++, which extends the consideration of local operations beyond mathematical calculations and incorporates more explicit constraints on visual perception and knowledge scope to automatically manage the reasoning process in the prompts. As shown in Tab.~\ref{exp:main-exp-2}, MARP++ achieves a 5\% improvement in accuracy, outperforming baselines in nearly all domains in M3CoT. See Appendix~D for more detailed prompting.

\subsection{Reasoning Boundary Scaling for Multimodal LLM}
In addition, to extend our findings to multi-modal scenarios, we tested M3CoT across three domains using five common multi-modal large models. As shown in Fig.~\ref{fig:correlation-multimodal} (a, b), a strong linear positive correlation between reasoning boundaries on the benchmark and model accuracy is observed in all three domains. However, as shown in Fig.~\ref{fig:correlation-multimodal} (b), the improvement in CFRB lags significantly behind CIRB in magnitude, similar to the text modality. This highlights a key issue with current multi-modal LLMs: they need to focus on enhancing the CFRB, an area that has yet to be fully mastered by the models.

	\section{Related Work}
This section reviews recent mechanism literature on Chain-of-Thought (CoT) prompting, with a focus on both theoretical and empirical studies~\cite{chen2025towards}. Previous work~\cite{madaan-etal-2023-makes,wang-etal-2023-towards,saparov2023language,he2023solving,zhang2024pattern,wang2024rethinking,prystawski2024think} demonstrates that LLMs learn reasoning chains through contextual demonstrations.
Lampinen \textit{et al.}~\cite{lampinen-etal-2022-language} and Tan \textit{et al.}~\cite{tan-2023-causal} establish a causal link between intermediate steps and final answers in qualitative experiments. Wang \textit{et al.}~\cite{wang2023large}, Hanna \textit{et al.}~\cite{hanna2024does} and Dutta \textit{et al.}~\cite{dutta2024think} explore LLM neural substructures, providing a white-box perspective on CoT reasoning and showing that LLMs generate multiple parallel answer paths.
Recently, a large amount of work has demonstrated the upper-bounds and limitations of LLM in various CoT tasks~\cite{qin2023cross,huang2024far,chen2025ecm}.
Bi \textit{et al.}~\cite{bi2024program} investigate these bounds on planning capability in code generation by training LLM on CoT samples of varying difficulties. Their findings suggest that LLMs have a limited capacity to learn or manage tasks exceeding a certain complexity threshold.
Further understanding of the CoT upper-bound, Merrill \textit{et al.}~\cite{merrill2023expressive}, and Feng \textit{et al.}~\cite{feng2024towards} analyze single-step arithmetic capability, which suggests an upper bound on model performance related to input length in single-step reasoning processes~\cite{li2023chain}.

While progress has been made in CoT explanations for LLMs, challenges persist, particularly the lack of quantifiable metrics for CoT's upper bounds and the absence of optimization guidelines. To address these issues, we introduce the reasoning boundaries framework++ (\texttt{RBF++}), which quantifies both measurable and unmeasurable reasoning boundaries. This framework also optimizes CoT approaches and extends theory to multimodal and reasoning LLM scenarios. The methodology offers a transferable, user-friendly tool for enhancing model performance from a mechanistic perspective, providing insights to guide future research and development.

\section{Conclusion}
In conclusion, this study introduces the Reasoning Boundary Framework++ (\texttt{RBF++}), which addresses two key challenges in Chain-of-Thought (CoT) reasoning for large language models (LLMs): the lack of metrics and guidance for boundaries with measurable  capabilities; the absence of metrics for boundaries with unmeasurable capabilities. By defining reasoning boundaries (RBs), their combination law, constant assumption, and reasoning boundary division mechanism, \texttt{RBF++} provides a clear method for assessing CoT performance and optimizing reasoning paths both in capability-measurable  and capability-unmeasurable scenarios across diverse models and tasks, including multimodal scenarios and reasoning LLMs, and offers valuable insights for improving CoT strategies. This work advances our understanding of reasoning in LLMs and provides tools for future optimization.
\newpage

	\bibliographystyle{IEEEtran}
	\bibliography{IEEEabrv,ref}

\begin{IEEEbiography}[{\includegraphics[width=1in,height=1.25in,clip,keepaspectratio]{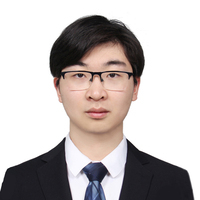}}]{Qiguang Chen} is currently a PhD student in Harbin Institute of Technology (HIT). He is a member of the Language Analysis Reasoning Group (LARG) within the Research Center for Social Computing and Interactive Robotics, Harbin Institute of Technology (SCIR). His research interests include mechanism modeling of chain-of-thought reasoning in large language models. He has published more than 10 papers in top journals and conferences, including NeuIPS, ACL, and EMNLP, etc.
\end{IEEEbiography}
	
	\vfill
\begin{IEEEbiography}[{\includegraphics[width=1in,height=1.25in,clip,keepaspectratio]{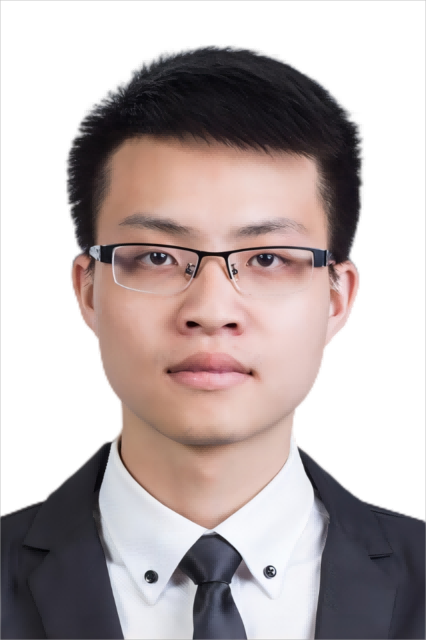}}]{Libo Qin} received his Ph.D. degree in computer science from the Harbin Institute of Technology (HIT), China. He is a Professor in School of Computer Science and Engineering, Central South University (CSU). His current research interests include reasoning, natural language processing and dialogue systems.
\end{IEEEbiography}
	
\vfill

\begin{IEEEbiography}[{\includegraphics[width=1in,height=1.25in,clip,keepaspectratio]{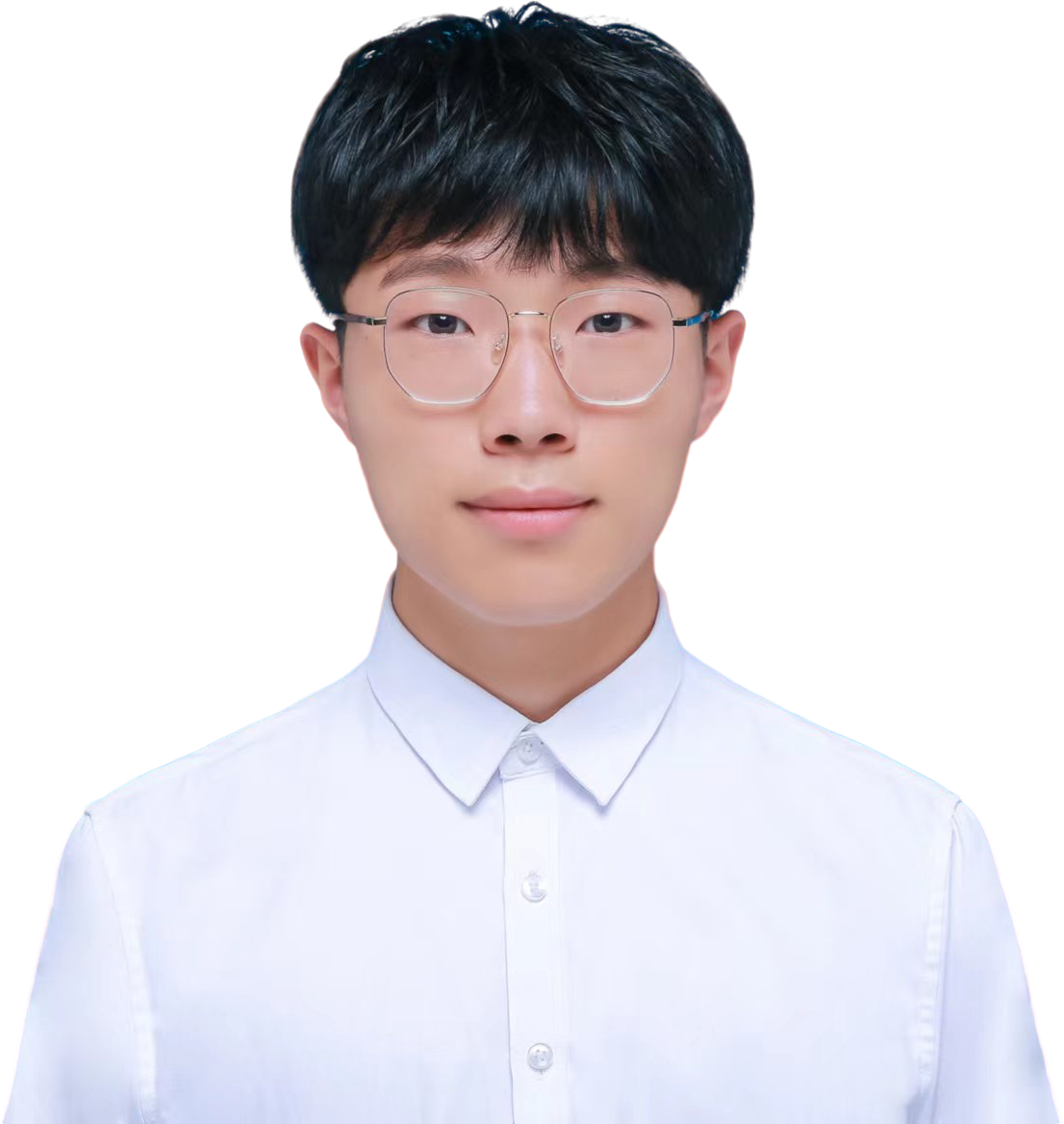}}]{Jinhao Liu} is currently an undergraduate student at the School of Artificial Intelligence, Harbin Institute of Technology (HIT). He is a member of the Language Analysis Reasoning Group (LARG) within the Research Center for Social Computing and Interactive Robotics, Harbin Institute of Technology (SCIR). His research interests include mechanism modeling of chain-of-thought reasoning in large language models.
\end{IEEEbiography}
	
\vfill

\begin{IEEEbiography}[{\includegraphics[width=1in,height=1.25in,clip,keepaspectratio]{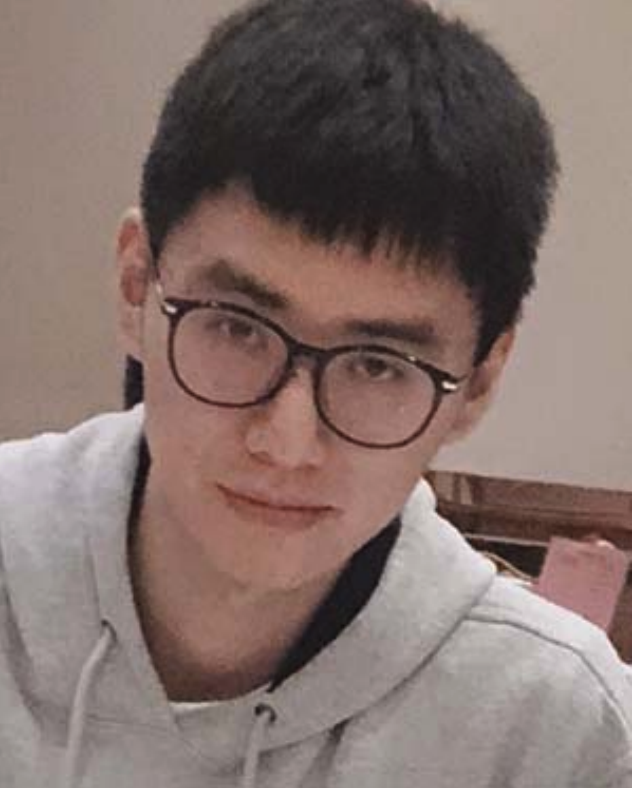}}]{Yue Liao} received the PhD degree from the School of Computer Science and Engineering, Beihang University. He is currently a postdoctoral fellow with MultiMedia Lab (MMLab), Chinese University of Hong Kong (CUHK) and Centre for Perceptual and Interactive Intelligence (CPII). His research interests include human-object interaction detection, multi-modality understanding and object detection. He has published more than 10 papers at top journals and conferences, including IEEE Transactions on Pattern Analysis and Machine Intelligence, IEEE Transactions on Image Processing, NeuIPS, CVPR, ICCV and ECCV, etc.
\end{IEEEbiography}
	
\vfill

\begin{IEEEbiography}[{\includegraphics[width=1in,height=1.25in,clip,keepaspectratio]{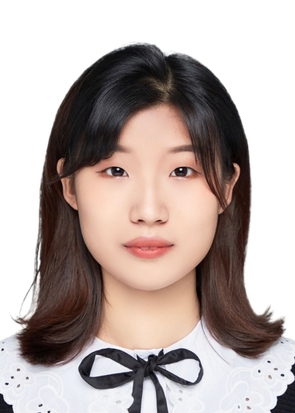}}]{Jiaqi Wang} is currently a second year PhD student at the Chinese University of Hong Kong.  Before that, she received her bachelor's degree from the Harbin Institute of Technology.  Her research interests include machine learning, agents, and large language models. She has published many papers at the top journals and conferences, including Transactions on Machine Learning Research, AAAI, NeurIPS, etc.
\end{IEEEbiography}
	
	\vfill
\begin{IEEEbiography}[{\includegraphics[width=1in,height=1.25in,clip,keepaspectratio]{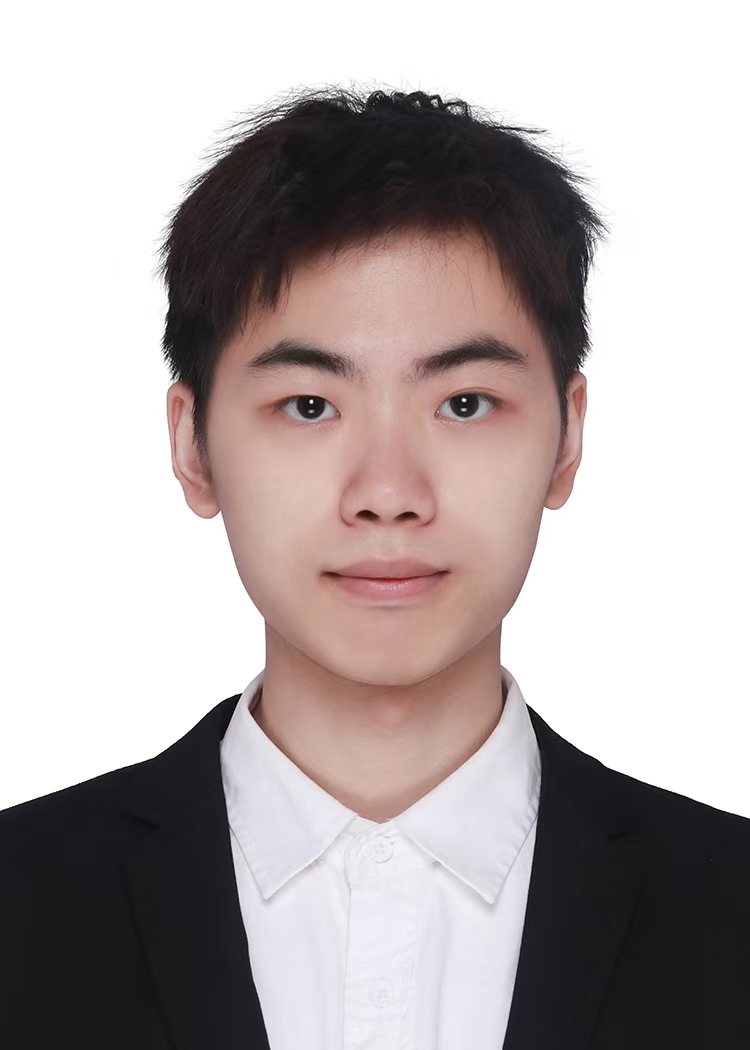}}]{Jingxuan Zhou}  is currently a second-year master’s student at the School of Computer Science and Engineering, Central South University. His research interests include tool learning and Chain-of-Thought in LLMs.
\end{IEEEbiography}
	
\vfill
\begin{IEEEbiography}[{\includegraphics[width=1in,height=1.25in,clip,keepaspectratio]{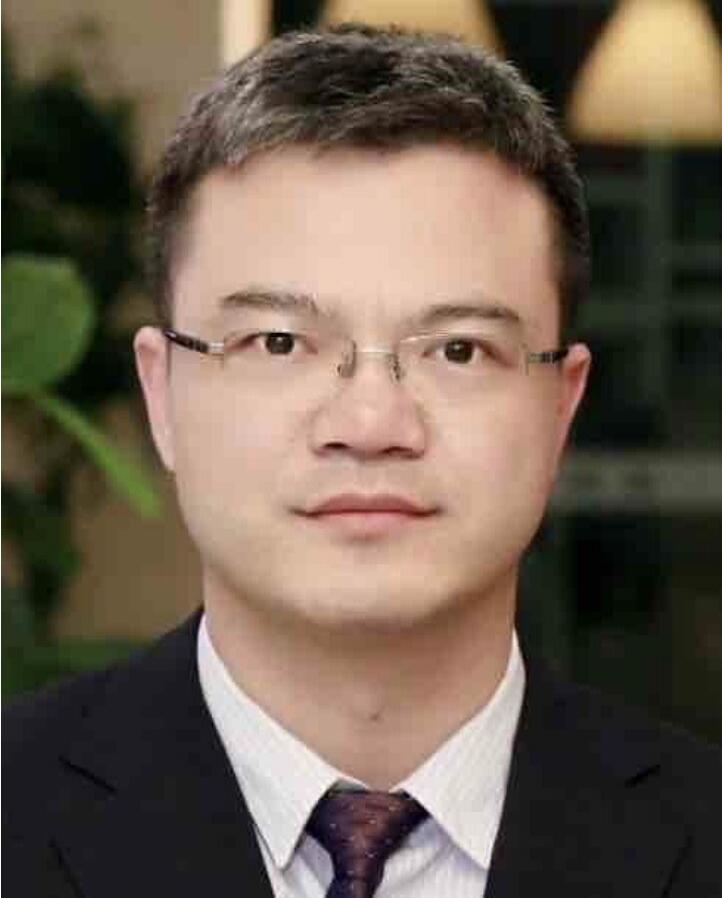}}]{Wanxiang Che} received his Ph.D. degree in computer science from the Harbin Institute of Technology (HIT), China, in 2008. He is a Full Professor at the Research Center for Social Computing and Interactive Robotics, School of Computer Science and Technology, Harbin Institute of Technology. His current research interests include natural language processing.
\end{IEEEbiography}
	
\vfill

\section*{Appendix}
\setcounter{subsection}{0}
\renewcommand{\thesubsection}{\Alph{subsection}}
\subsection{Mathematical Analysis \& Proof}
\label{append:proof}
\subsubsection{Definitions \& Assumptions}

In order to further quantify and analyze the combination law of RB, we will define the concept of difficulties for different tasks:
\begin{definition}
	The difficulty of solving a certain problem during model reasoning is an independent constant.
\end{definition}

That is, the difficulty $\mathcal{D} (t_1, t_2)$ satisfies:
\begin{equation}
	\mathcal{D} (t_1, t_2|m) = \mathcal{D} (t_1|m) + \mathcal{D} (t_2|m) = K_1 + K_2,
\end{equation}
where, $K_1, K_2$ denotes the relevant constants.
Therefore, the combined difficulty formally satisfies:
\begin{align*}
	\mathcal{D} &(t_1, t_2, \dots ,t_n|m)\\
	&= \mathcal{D} (t_1, t_2, \dots, t_{i-1}, t_{i+1}, \dots, t_n|m) + \mathcal{D} (t_i|m)\\
	&= \sum_i \mathcal{D}(t_i|m)
\end{align*}

\begin{definition}
	The RB is defined as the reciprocal of the difficulty of solving the problem. The greater the difficulty of solving the problem, the lower the RB and the smaller the feasible area.
\end{definition}

Therefore, the combination law of RB satisfies:
\begin{equation}
	\mathcal{B} (t_1, t_2, \dots ,t_n|m) \propto \frac{1}{\mathcal{D} (t_1, t_2, \dots ,t_n|m)}\label{eq:proof-0}
\end{equation}

\begin{definition}
	If all basic RBs are infinite, it means that all the difficulties approach to the zero and the model is omnipotent. Therefore, the combined RB is also infinite.
	\label{def:all-inf}
\end{definition}
 Formally, the combination law satisfies that: 
\begin{equation}
	\mathcal{B}(+\infty, +\infty, \dots, +\infty|m) = +\infty
\end{equation} 
\begin{assumption}
	The combination law function is continuously differentiable everywhere.
	\label{asp:1}
\end{assumption}
\begin{assumption}
	All basic reasoning boundary for combined reasoning boundary are mutually independent.
	\label{asp:2}
\end{assumption}
\subsubsection{The Proof of Combination Law}
Based on the above definitions and assumptions, we need to prove that the combination law is a combined RB and is the weighted harmonic average of two basic RBs.

\textbf{\textit{Proof.} }
Following Eq.~\eqref{eq:proof-0}, $\mathcal{D}(x_1, x_2, \dots, x_n|m)$ can be defined as:
\begin{equation}
	\mathcal{D}(x_1, x_2, \dots, x_n|m) = \sum^{n}_{i=1} \mathcal{D}(0,\dots, x_i,\dots,0|m).
\end{equation}

According to the Taylor expansion formula, we expand this formula at $x_i \rightarrow k_i$, we can get:
\begin{align}
	\mathcal{D}(x_1, x_2, \dots, x_n|m)&=\sum^{n}_{i=1}\sum_{j=1}^{+\infty}N_{ij}(x_{i}-k_{i})^{j}\\
	&=\sum^{n}_{i=1}N_{i1} (x_{i}-k_{i}) + \mathcal{O}(x_i) \\
	& \approx \sum^{n}_{i=1}N_{i1} (x_{i}-k_{i}),
\end{align}
where $N_{i1}=\frac{\partial \mathcal{D}(x_1, x_2, \dots, x_n|m)}{\partial x_i}$.
We set $t_i = \frac{1}{x_i}+b_i$ and $\frac{1}{\mathcal{B}(t_1, t_2, \dots, t_n|m)} \propto \mathcal{D}(x_1, x_2, \dots, x_n|m)$. Then the original formula is expressed as:
\begin{align}
	\mathcal{B}(t_1, t_2, \dots, t_n|m) &\approx \frac{N_0}{\sum^{n}_{i=1}\frac{N_{i1}}{t_i-b_i}-k_i} +N_{i1} k_0 \\
	&= \frac{1}{\sum^{n}_{i=1}\frac{N^{'}_{i1}}{t_i-b_i}-k^{'}_i}+k_0, \label{eq:combine-proof}
\end{align}
where $t_i$ represents the specific task measurement value, and $N_0$ and $k_0$ denote the linear parameters.
Given the minimal change in the derivative within the observable range, $N^{'}_{i1}=\frac{N_{i1}}{N_0}$ is treated as a constant $N_{i}$ in this task for simplicity. Experimental results show that, if sub-RBs are separated independently, $k^{'}_i=\frac{k_{i}}{N_0}$ and $k_0$ is typically 0. Since $t_i$ cannot be directly quantified, we use basic form of $\mathcal{B}(t_i|m)$ as its quantized substitute, thus simplifying the combination law as:
\begin{equation}
	\mathcal{B}(t_1, t_2, \dots, t_n|m) \approx \frac{1}{\sum^{n}_{i=1}\frac{N_{i}}{\mathcal{B}(t_i|m)-b_i}}. \label{eq:combine-final}
\end{equation}

\subsection{Details of Dataset}
\label{append:data-cons}
To assess the reasoning boundaries of LLMs effectively, it is crucial to develop a dataset that captures a range of complexities. To address this, we propose a novel approach for constructing a mathematical reasoning dataset through manual synthesis and annotation, resulting in the \textsc{BigGSM} and \textsc{BigGSM++} benchmarks. Our method involves the following steps:
\subsubsection{Domain Template Generation} We begin by using a prompt-driven LLM (GPT-4 and GPT-4o) to generate complex scenarios that require multi-step calculations, alongside initial example templates. The prompt provided to the model is:
\begin{prompt}
	Generate a scenario-related template involving multiple mathematical steps to solve a real-world problem. Ensure the scenario requires the application of different mathematical concepts. Please use "[VAR]" as a variable to mark the template of the question.
\end{prompt}
\subsubsection{Natural Language Template Creation} 
Recognizing that LLMs can produce errors and logical inconsistencies, we refine these initial templates to improve their accuracy and add mathematical calculations. To facilitate the generation of extended sequences, we decompose the templates into smaller, loopable segments that incrementally meet the multi-step reasoning demands.

\subsubsection{Domain Template Augmentation} 
To overcome the limited diversity of individual samples and better assess LLMs' mathematical capabilities, we use GPT-4 to generate at least three alternative augmented templates for each original template and step.
The generation prompt is:
\begin{prompt}
	Create three alternative versions of the following template that introduce different complexities or variables, ensuring each version demands an equivalent level of reasoning.
\end{prompt}

\begin{figure*}[t]
	\centering
	\includegraphics[width=0.98\textwidth]{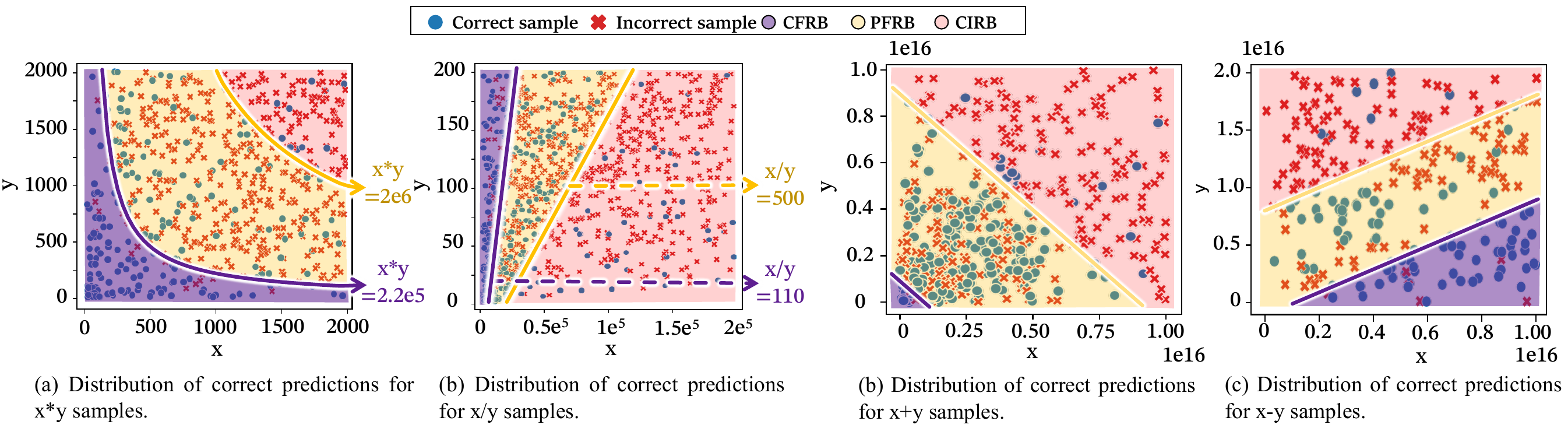}
	\caption{
		Existence verification for reasoning boundaries on basic arithmetic calculation tasks, including division, addition, and subtraction operations.
	}
	\label{fig:atom-rg-2}
\end{figure*}
\begin{table}[t]
	\centering
	\caption{
		Model list. In order to ensure a certain ability to follow instructions, we use the Instruct version of the model as much as possible (if available).
	}
	\begin{adjustbox}{width=0.80\linewidth}
		\begin{tabular}{lll}
			\toprule
			Model & Base Model & Parameters (B) 
			\\
			\midrule
			\rowcolor{gray!8}\multicolumn{3}{c}{\textit{Open-source General LLM}}\\
			\midrule
			LLaMA~\cite{touvron2023llama} & - & 7, 13, 33, 65 \\
			LLaMA-2~\cite{touvron2023llama2} & - & 7, 13, 70 \\
			LLaMA-3~\cite{touvron2023llama3} & - & 8, 70 \\
			Code-LLaMA~\cite{roziere2023code} & LLaMA-2~\cite{touvron2023llama2} & 7, 13, 34, 70 \\
			Mistral~\cite{jiang2023mistral} & - & 7 \\
			\midrule
			\rowcolor{gray!8}\multicolumn{3}{c}{\textit{Close-source General LLM}}\\
			\midrule
			Gemini-1.0-Pro~\cite{team2023gemini} & - & - \\
			GPT3.5-Turbo~\cite{openai2022gpt35} & -  & - \\
			Claude-3-Haiku~\cite{anthropic2024claude3} & -  & - \\
			Claude-3-Sonnet~\cite{anthropic2024claude3} & -  & - \\
			Claude-3-Opus~\cite{anthropic2024claude3} & -  & - \\
			GPT4o-mini~\cite{achiam2023gpt} & -  & - \\
			GPT4~\cite{achiam2023gpt} & -  & - \\
			GPT4o~\cite{achiam2023gpt} & -  & - \\
			\midrule
			\rowcolor{gray!8}\multicolumn{3}{c}{\textit{Open-source Math LLM}}\\
			\midrule
			MAmmoTH~\cite{yue2023mammoth} & LLaMA-2~\cite{touvron2023llama2}  & 7,13 \\
			MAmmoTH~\cite{yue2023mammoth} & Mistral~\cite{jiang2023mistral}  & 7 \\
			OpenMATH-Instruct~\cite{toshniwal2024openmathinstruct} & LLaMA-2~\cite{touvron2023llama2} & 70 \\
			OpenMATH-Instruct~\cite{toshniwal2024openmathinstruct} & Mistral~\cite{jiang2023mistral}  & 7 \\
			\midrule
			\rowcolor{gray!8}\multicolumn{3}{c}{\textit{Multimodal LLM}}\\
			\midrule
			DeepSeek-v3~\cite{guo2025deepseek} & -  & 671 \\
			DeepSeek-R1~\cite{guo2025deepseek} & DeepSeek-v3~\cite{guo2025deepseek}  & 671 \\
			o3-mini-low & - & - \\
			o3-mini-medium & - & - \\
			o3-mini-high & - & - \\

			\midrule
			\rowcolor{gray!8}\multicolumn{3}{c}{\textit{Multimodal LLM}}\\
			\midrule
			Qwen-VL-Max & -  & - \\
			Qwen-VL-Plus & -  & - \\
			GPT4o-mini~\cite{achiam2023gpt} & -  & - \\
			GPT4o~\cite{achiam2023gpt} & -  & - \\
			Gemini-1.5-Pro & -  & - \\
			Gemini-2.0-Flash-Thinking & -  & - \\
			\bottomrule
		\end{tabular}
	\end{adjustbox}
	
	\label{exp:model}
\end{table}

\subsubsection{Numeric Filling} 
After preparing the templates, we test the LLMs' computational reasoning limits by introducing numerical values ranging from 1 to 1e5 in multiplication tasks. This step evaluates the models' performance across a broad range of numerical challenges.
\subsubsection{Manual Annotation} 
To ensure the accuracy and logical coherence of our synthetic samples, we manually review them to correct any errors from the automated generation process.
Three experts were then hired to assess whether the samples were correct. Only those samples with consensus from at least two experts were retained. The Cohen’s kappa value for expert agreement was 0.92, indicating near-perfect consistency.

\subsection{Analysis for Complex-CoT and Least-to-Most}
\label{append:analysis}

\paragraph{Complex CoT (CCoT) Prompting can achieve better CoT in textual scenarios within a specific RB by simplifying the calculation reasoning step.}
We believe that CCoT optimizes the performance of the model by allowing the model to reach its computational limit as much as possible in single-step reasoning.
Therefore, the combined RB for CCoT can be expressed as:
\begin{equation}
	\mathcal{B}^{\texttt{CCoT}}(p, c) = \lim\limits_{\mathcal{B}(c) \rightarrow \mathcal{B}_{\text{Acc}=100\%}(c)}\frac{1}{\frac{N_{1}}{(\mathcal{B}(c)-b_1)} + \frac{N_{2}}{(\mathcal{B}'(p)-b_2)}}
\end{equation}
Assuming the premises of RB remain unchanged (${\mathcal{B}}^{\texttt{Complex}}(p, c)=\mathcal{B}^{\texttt{CoT}}(p, c)$), it can obviously yield the solution $\mathcal{B}'(p)>\mathcal{B}(p)$. Therefore, the model can accept more steps of reasoning boundary, that is, if the planning difficulty $d_p$ is less than reasoning capability $\mathcal{B}'(p)$, the accuracy is higher.

\paragraph{Least-to-Most Prompting can achieve better CoT in textual scenarios within a specific RB by simplifying the planning reasoning paths.}
Least-to-most prompting structures problem-solving hierarchically, by breaking questions into smaller sub-questions and further solving them one-by-one.
Accordingly, the Least-to-most RB can be divided into three sub-RBs, namely, the problem decomposition RB $\mathcal{B}(d)$, the problem planning RB $\mathcal{B}(p)$, and the single-step calculation RB $\mathcal{B}(c)$.
Therefore, the combined RB for least-to-most can be expressed as:
\begin{equation}
	\mathcal{B}^{\texttt{LtM}}(d, p, c) = \frac{1}{\frac{N_{1}}{(\mathcal{B}'(c)-b_1)} + \frac{N_{2}}{(\mathcal{B}(p)-b_2)} + \frac{N_{3}}{(\mathcal{B}(d)-b_3)}}.\label{eq:LtM-origin}
\end{equation}
Ideally, if the problem decomposition ability of the model is excellent ($\mathcal{B}(d)\rightarrow +\infty$), it can decompose the problem into sub-problems that can be solved in one step every time $\mathcal{B}(c) \rightarrow 1$, therefore the least-to-most RB can be expressed as:
\begin{align}
	\hat{\mathcal{B}}^{\texttt{LtM}}(d, p, c) &= \lim\limits_{\mathcal{B}(c) \rightarrow 1, \mathcal{B}(d)\rightarrow +\infty}\mathcal{B}^{\texttt{LtM}}(d, p, c) \\
	&= \frac{\mathcal{B}'(c)-b_2}{N_{1}(\mathcal{B}'(c)-b_2)-N_{2}}, \label{eq:LtM}
\end{align}

Assuming the premises of RB remain unchanged ($\hat{\mathcal{B}}^{\texttt{LtM}}(d, p, c)=\mathcal{B}^{\texttt{CoT}}(p, c)$), it can obviously yield the solution $\mathcal{B}'(c)>\mathcal{B}(c)$. On the contrary, the model can accept larger difficulty $d$, which also shows that using least-to-most prompting can effectively increase the maximum of acceptable calculation RB under a given RB, thereby improving model performance.
However, the performance improvement of the model is not significant, which we attribute to the fact that the current model cannot push its performance to the ideal limit.

\subsection{MARP and MARP++ Implementation}
\label{append:heuristic-baselines}

\subsubsection{Minimum Acceptable Reasoning Paths (MARP)}
To address the limitations of previous CoT strategies, we propose Minimum Acceptable Reasoning Paths (MARP). Firstly, to reduce the model's computational load, we introduce instructions that limit its single-step computing power, thereby optimizing its reasoning boundary. Secondly, to enhance the model's acceptability, we increase the computation amount per step within this boundary and reduce the number of global planning steps, thus alleviating planning pressure.
\begin{prompt}
	You need to perform multi-step reasoning, with each step carrying out as many basic operations as possible.
	Remember, you can only complete tasks that contain up to 5 basic operations per step, and multiplication operations must be less than 1.5e5. The upper limit of the multiplication operations decreases as the number of operations per step increases.
\end{prompt}

\subsubsection{Minimum Acceptable Reasoning Paths++ (MARP++)}

\begin{prompt}
You are required to perform multi-step reasoning, ensuring that each step operates within clearly defined boundaries:
\begin{itemize}[leftmargin=4pt]
	\item Global Planning Boundary: Focus on the overall strategy and high-level goal. You should break down the task into manageable steps (less than 15 steps) within your capabilities but always consider the broader objective to ensure coherence in the approach.
	\item Local Step Operation Boundary: In each step, perform as many basic operations as possible, but each step must adhere to a limit of 5 basic operations. Avoid exceeding this boundary to maintain clarity and precision at each stage.
	\item Multimodal Perception Boundary: When reasoning, incorporate all available information (text, images, etc. if available) without overstepping the boundaries of what can be processed in one step. Make sure to integrate the relevant modalities effectively within the defined operation limits. If perception is very difficult, please divide it into multiple steps for multimodal perception.
	\item Multimodal Perception Boundary: Domain-Knowledge Boundary: Utilize your domain knowledge effectively but ensure that each step remains grounded within your expertise. Do not go beyond what is strictly necessary for the current step.
\end{itemize}
\end{prompt}

\end{document}